\documentclass{llncs}
\usepackage{amssymb}
\usepackage{multirow}
\usepackage{color}
\usepackage{graphicx}
\newcommand{\argmin}{\arg\!\min}
\newcommand{\bm}[1]{ \mathbf #1} 
\begin{document}
\frontmatter          % for the preliminaries
\pagestyle{empty}  % switches off printing of running heads
\mainmatter              % start of your contributions
\title{Towards ultra-high resolution 3D reconstruction of a whole rat brain from 3D-PLI data}
%\title{Towards meso- and micro-scale connectome reconstruction of RAT brain using 3D-PLI data}
%
%\titlerunning{Short Title}  % abbreviated title (for running head)
%
%
%
%\author{Rasmus Larsen\inst{1} \and author2\inst{1,2}  \and author3\inst{3} }
%
%\authorrunning{R. Larsen et al.}   % abbreviated author list (for running head)
%
%%%% modified list of authors for the TOC (add the affiliations)
%\tocauthor{Rasmus Larsen (Technical University of Denmark),
%author2 (Affiliation of Author2), author3 (Affiliation of
%Author3), }
%
%\institute{Informatics and Mathematical Modelling, Technical
%University of Denmark, Denmark, \email{rl@imm.dtu.dk}
%  \and
%  Department of Radiology, University Hospital, Denmark,
%  \and
% Innovative Solutions Corp., USA\thanks{We are thankful to XYZ}
%}

% The following lines are used to remove authors and affiliations from
% the front page and running titles
\author{Sharib Ali$^1$, Martin Schober$^2$, Philipp~Schl{\"o}me$^{2}$, Katrin~Amunts$^{2,3}$, Markus Axer$^2$, \and Karl Rohr$^1$}
\authorrunning{Ali et al.}
\titlerunning{3D reconstruction of a whole rat brain}
\institute{$^1$Dept. of Bioinformatics, Biomedical Computer Vision Group, BIOQUANT, IPMB, DKFZ,  University of Heidelberg, Germany\\
$^2$Institute of Neuroscience and Medicine~1, Research Centre J\"{u}lich, J\"{u}lich, Germany\\ $^3$C{\'e}cile and Oskar Vogt Institute of Brain Research, Heinrich Heine University D{\"u}sseldorf, University Hospital D{\"u}sseldorf, Germany}
%
%\email{sharib.ali@eng.ox.ac.uk}
\maketitle           
\begin{abstract}
3D reconstruction of the fiber connectivity of the rat brain at microscopic scale enables gaining detailed insight about the complex structural organization of the brain. We introduce a new method for registration and 3D reconstruction of high- and ultra-high resolution (\,64 $\mu$m and 1.3 $\mu$m pixel size) histological images of a Wistar rat brain acquired by 3D polarized light imaging (3D-PLI). Our method exploits multi-scale and multi-modal 3D-PLI data up to cellular resolution. We propose a new feature transform-based similarity measure and a weighted regularization scheme for accurate and robust non-rigid registration. To transform the 1.3 $\mu$m ultra-high resolution data to the reference blockface images a feature-based registration method followed by a non-rigid registration is proposed. Our approach has been successfully applied to 278 histological sections of a rat brain and the performance has been quantitatively evaluated using manually placed landmarks by an expert.
%\keywords{3D-PLI brain images, Multi-modal image registration, feature transform}
\end{abstract}
\section{Introduction}
\label{sec:intro}
Studying the brain fiber architecture and their functionality, like that of the rat brain, is important for understanding complex human brain organization. Conventional imaging methods include electron microscopy (EM), optical microscopy (OM), and diffusion magnetic resonance imaging (D-MRI). While D-MRI is limited in resolution, EM and OM often require some selective staining procedure of histological brain sections to reveal fiber connectivity. Recent advances in 3D polarized light imaging (3D-PLI, a specialized OM technique that utilizes the birefringence of nerve fibers) allows acquiring high- and ultra-high resolution images of fibrous brain tissues  \cite{AxerET11:NeuroImage}. In addition, information about 3D fiber orientation can be obtained without staining. 3D-PLI data consists of different image modalities (Fig. 1, right): Transmittance map representing the extinction of polarized light when passing through the brain tissue, Retardation map showing the tissue's (fiber's) birefringence, as well as direction and inclination maps representing the local 3D fiber orientation. Blockface images are acquired during the sectioning procedure (Fig. 1, left) and constitute undistorted reference images for the acquired histological sections. 
\begin{figure*}[t!]
\fbox{
\begin{minipage}[b]{0.3\linewidth}
  \centering
  \centerline{\includegraphics[trim=0cm 0cm 1cm 0cm, clip=true, scale = 0.152, angle =90]{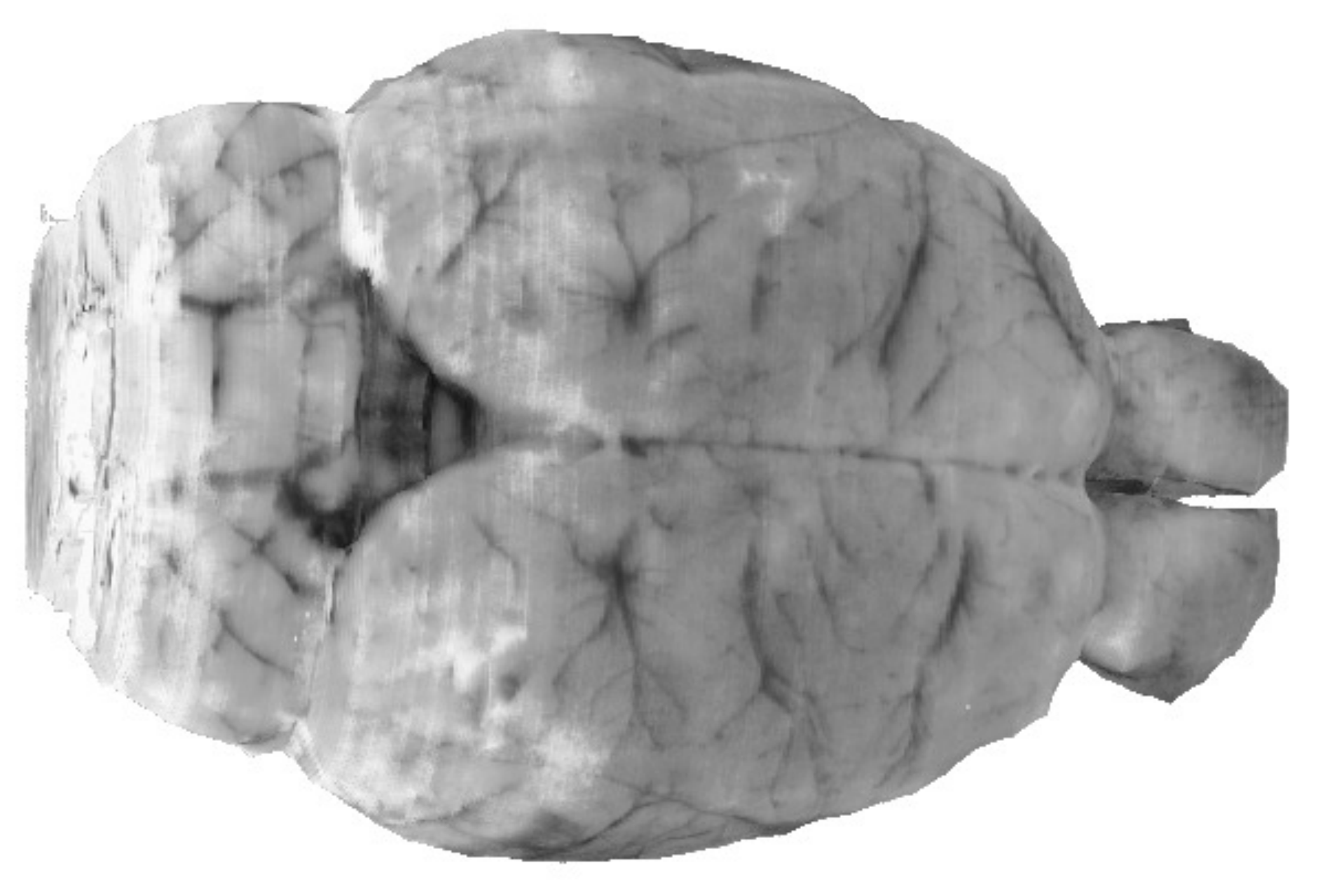}}
  \centerline{\includegraphics[trim=0cm 1cm 1cm 0cm, clip=true, scale = 0.024]{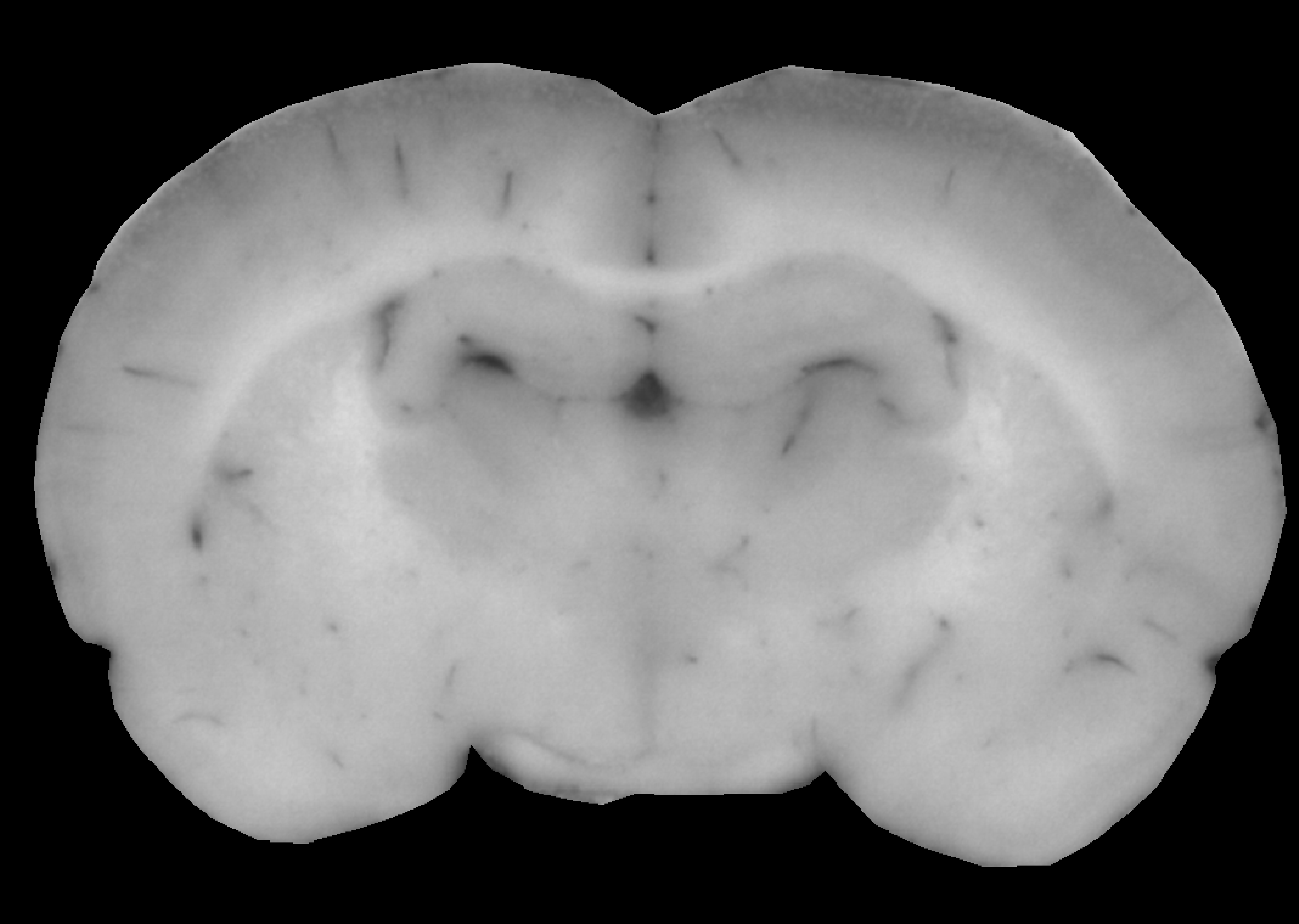}}
\end{minipage}
}
\fbox{
\begin{minipage}[b]{0.3\linewidth}
  \centering
  \centerline{\includegraphics[trim=0cm 2cm 0cm 2cm, clip=true, height=3.15cm, width =2.0cm, angle=90]{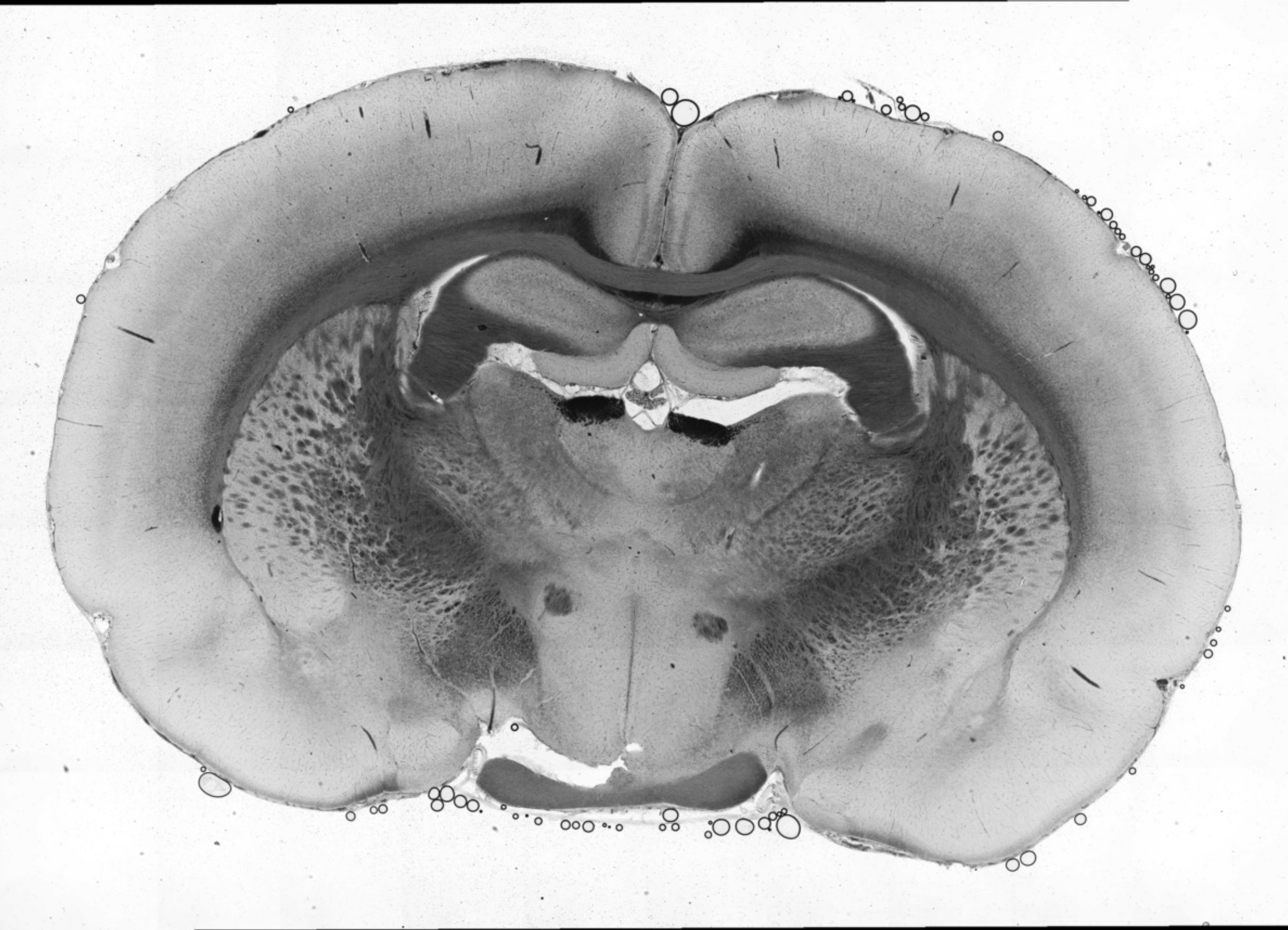}}
  \centerline{\includegraphics[trim=0cm 1cm 0cm 2cm, clip=true, height=3.15cm, width =1.9cm, angle=90]{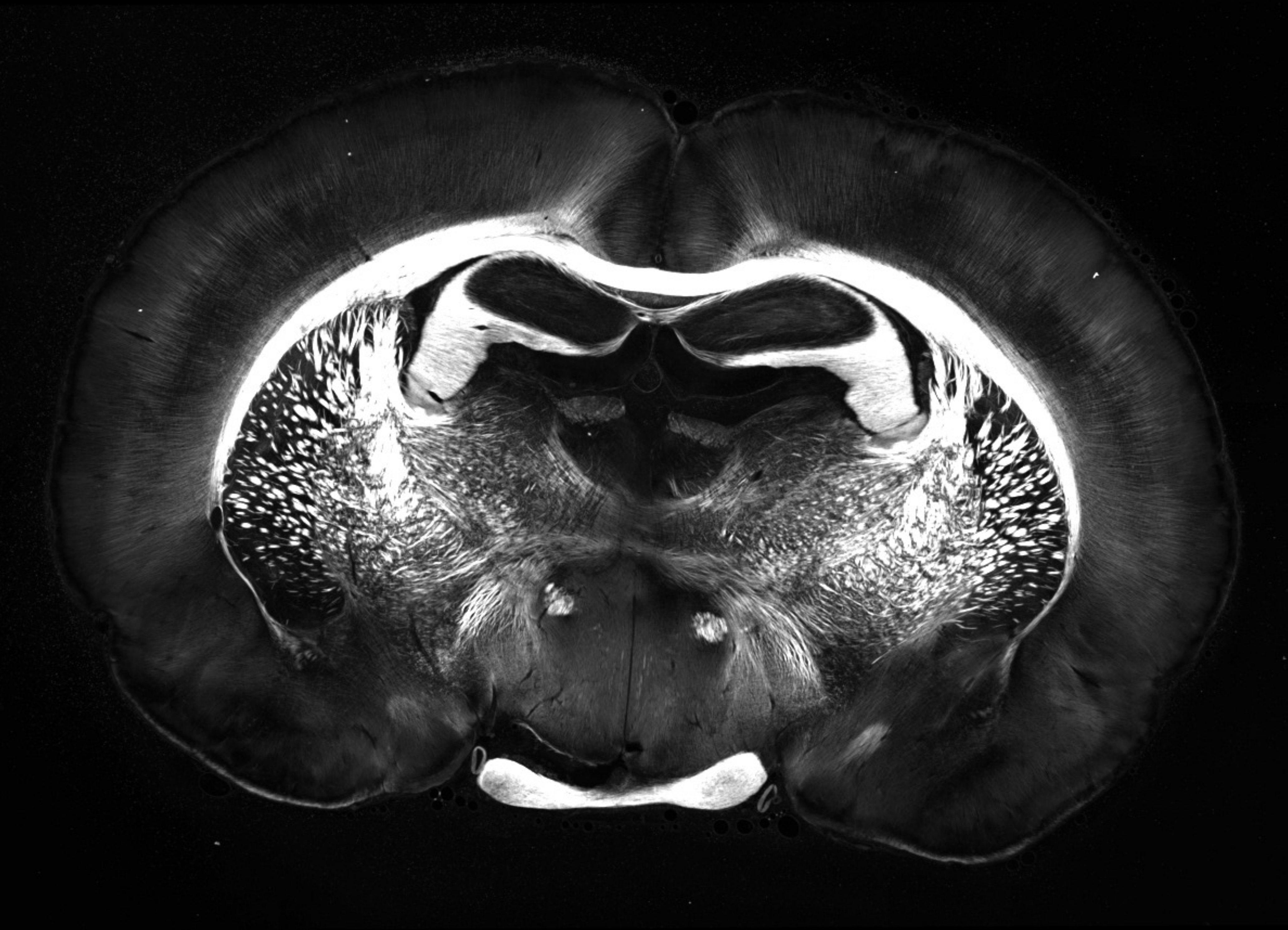}}
  \end{minipage}
  \begin{minipage}[b]{0.32\linewidth}
  \centerline{\includegraphics[trim=0cm 1cm 0cm 2cm, clip=true, height=3.15cm, width =1.9cm,angle=90]{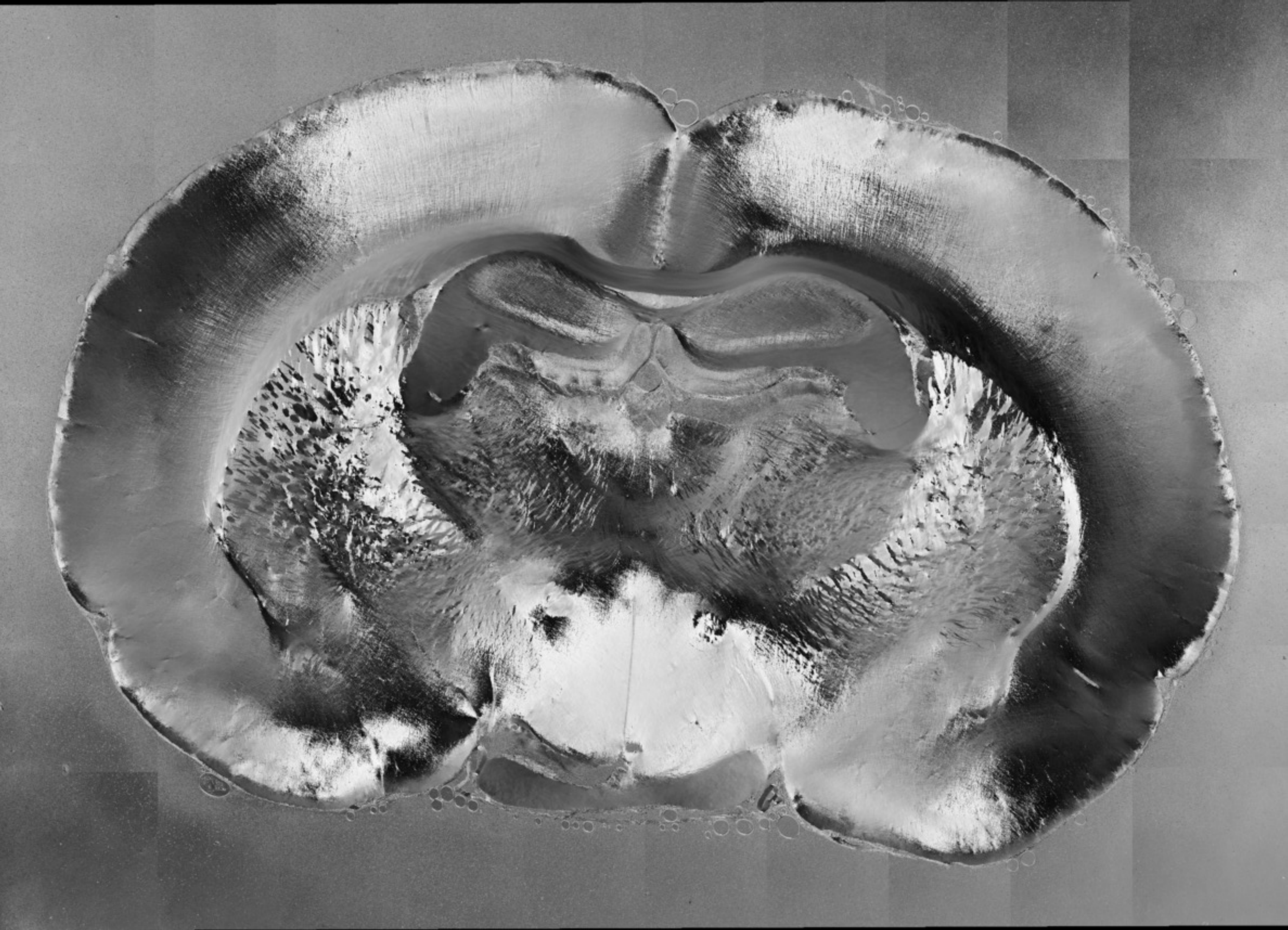}}
  \centerline{\includegraphics[trim=0cm 1cm 0cm 2cm, clip=true, height=3.15cm, width =1.9cm,angle=90]{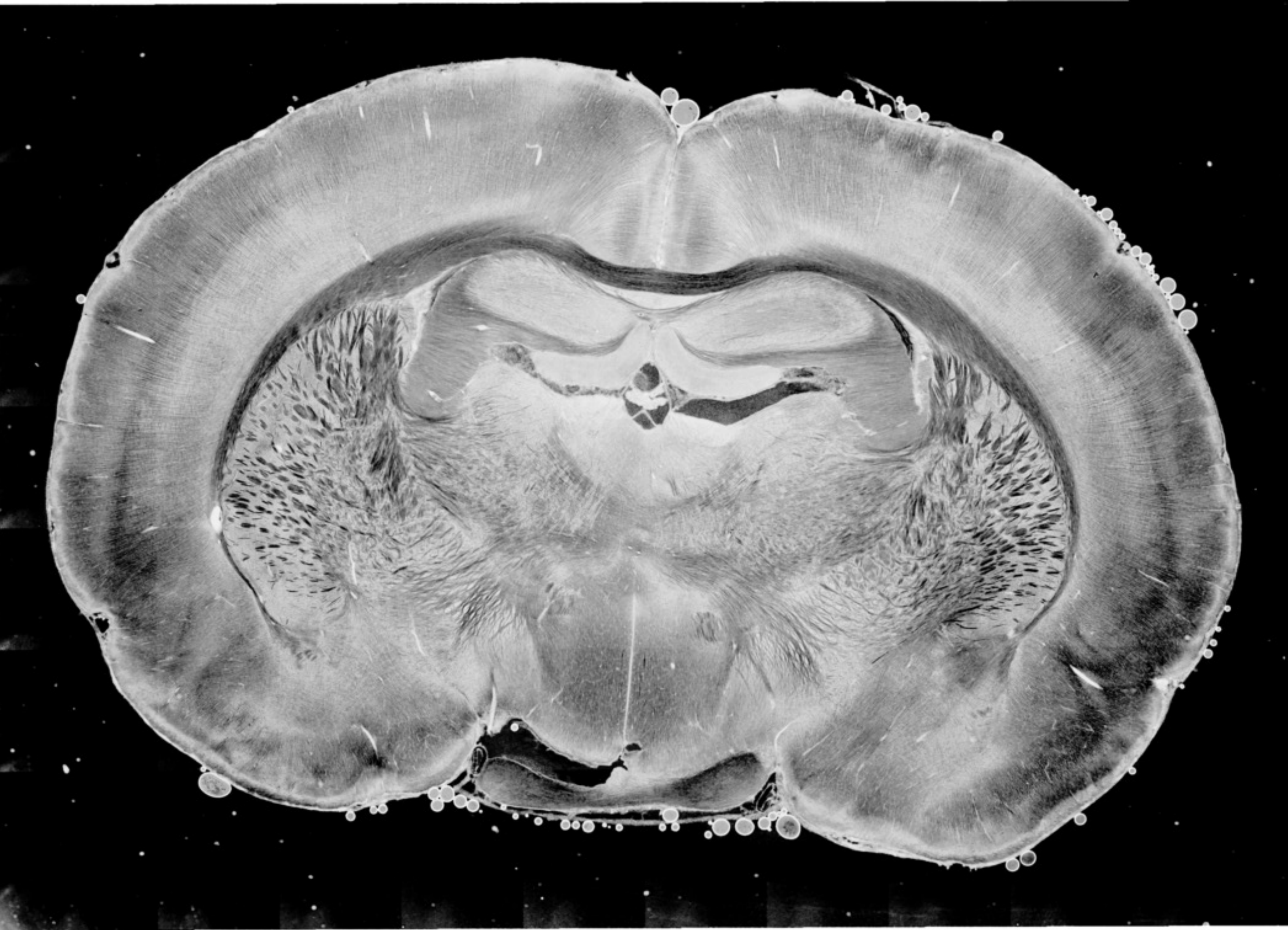}}
\end{minipage}}
\vspace{-0.5cm}
\caption{Rat brain data. Left box: Reference blockface with 3D blockface volume (top) and mid-section (bottom). Right box: original ultra-high resolution (1.3$\mu$m) 3D-PLI data comprising transmittance (top left), retardation (bottom left), direction (top-right), and inclination (bottom-right) maps. Images are scaled for better visualization.}
\label{fig:data}
\end{figure*}

During the sectioning and mounting process brain tissue undergoes strong distortions. Thus, spatial coherence between sections is lost and hence image registration becomes an inevitable task. In previous work, 3D reconstruction of histological sections of the rat brain~(e.g.,~\cite{LebenbergEt10:NI,Ourselin01,MajkaWojcik16:NI}) was performed using rigid or affine registration (e.g.~\cite{Ourselin01,Arsigny:MICCAI03}), which is generally not sufficient to cope with deformations in histological sections as mentioned in~\cite{LebenbergEt10:NI}. \cite{MajkaWojcik16:NI} used affine registration with subsequent diffeomorphic non-rigid registration employing mutual information. Compared to traditional histological data, 3D-PLI relies on unstained cryo-sections and is acquired at very different resolutions. This poses different challenges compared to traditional histological data. In previous work on the registration and 3D reconstruction of 3D-PLI data, high-resolution images ($64\mu$m pixel size) were used in~\cite{Ali1Et17:ISBI,SchubertET16:FIN} and ultra-high resolution~($1.3~\mu$m pixel size) images in \cite{Ali2Et17:ISBI}. However, in \cite{Ali2Et17:ISBI} only rigid registration of the ultra-high resolution data to unregistered high-resolution images was performed, and the human brain was considered but not rat brain.~\cite{Ali1Et17:ISBI} used high-resolution human brain sections (64~$\mu$m pixel size) for registration to reference blockface data of the same resolution. Note that human brain sections typically cover larger areas, contain more prominent structures, and include less image noise compared to the rat brain. Thus, registration of 3D-PLI data of the rat brain is more difficult. In \cite{SchubertET16:FIN}, high-resolution 3D-PLI data (64~$\mu$m) of the rat brain was first registered to the blockface data of same resolution and then transformed to a reference Waxholm space. However, in contrast to~\cite{SchubertET16:FIN}, we register high-resolution images with a section thickness of $60~\mu$m to blockface images of 15.5~$\mu$m pixel resolution, which is more challenging due to the large scale difference. Also, we subsequently register ultra-high resolution images ($1.3~\mu m$) first to the registered high-resolution images ($15.5~\mu$m after scaling) and then to upscaled reference blockface images at $1.3~\mu m$ resolution using non-rigid registration (each image section has a size of about $15000\times 12000$ pixels). In addition, whereas in~\cite{SchubertET16:FIN} B-splines and a fluid model were used, respectively, we here use a more realistic deformation model based on Gaussian non-rigid body splines (GEBS) for non-rigid registration. None of the previous work provided a complete framework for ultra-high resolution 3D reconstruction of the rat brain from 3D-PLI. 

In this contribution, we introduce a new method for multi-scale (both high- and ultra-high resolution data) and multi-modal registration of histological rat brain sections from 3D-PLI.
The main contributions are: 1) registration of 3D-PLI data with three different spatial resolutions~($1.3~\mu m$, $15.5~\mu m$, and $64~\mu m$ pixel size), 2) correlation transform-based similarity metric for efficient and robust rigid registration, 3) introduction of a feature transform-based similarity metric and weighted regularization for non-rigid registration using a physically-based deformation model, 4) robust feature-based registration, and 5) a complete pipeline for 3D reconstruction. 
\begin{figure}[t!]
\begin{minipage}[b]{0.34\linewidth}
  \centering
  %TODO:crop the image
  \centerline{\includegraphics[trim=0cm 1cm 0cm 2cm, clip=true, height=4.15cm, width =3.9cm,angle=0]{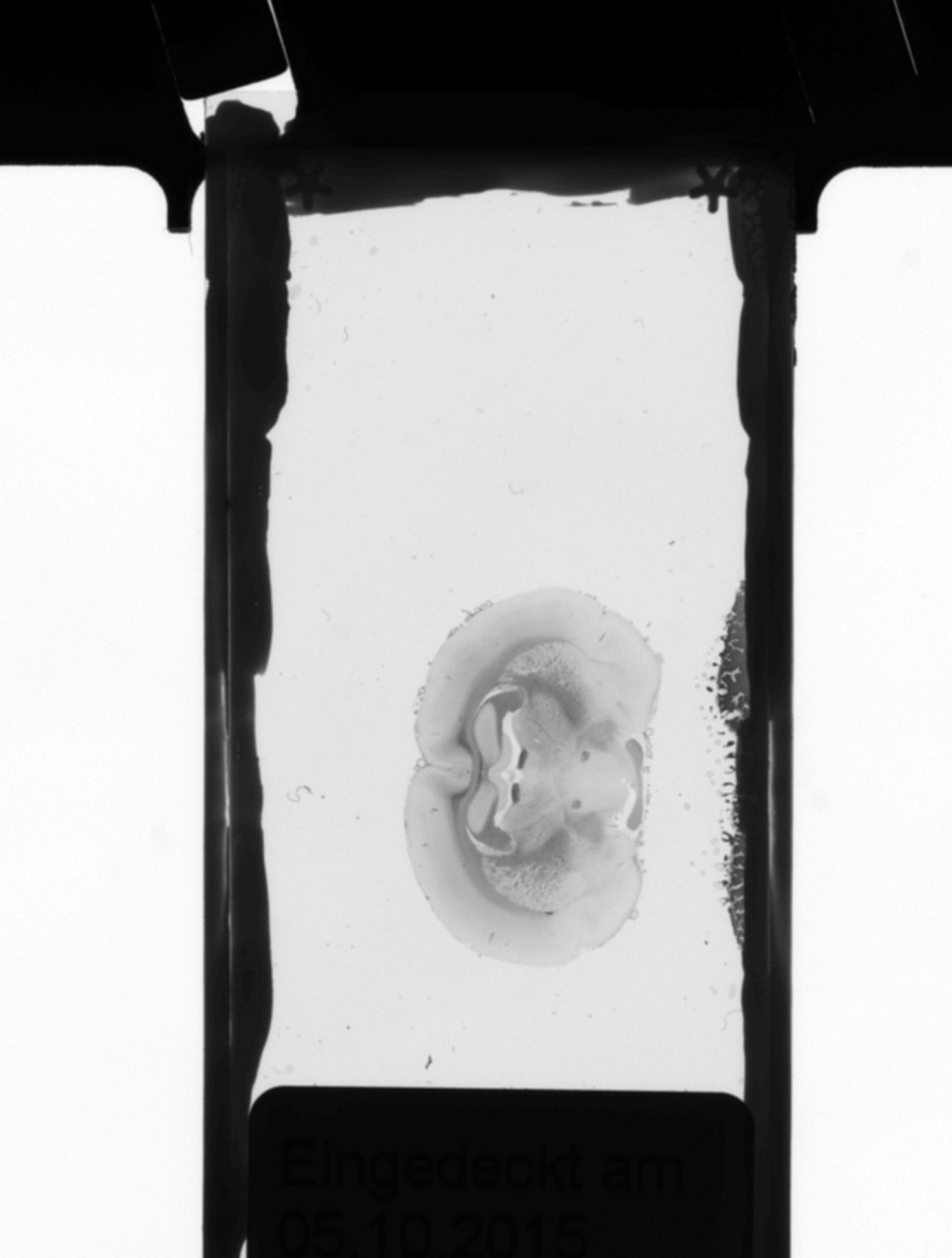}}
 % \centerline  {(a)}\medskip
\end{minipage}
\begin{minipage}[b]{0.32\linewidth}
  \centering
  %TODO:crop the image
  \centerline{\includegraphics[trim=0cm 0cm 0cm 0cm, clip=true,scale=0.70]{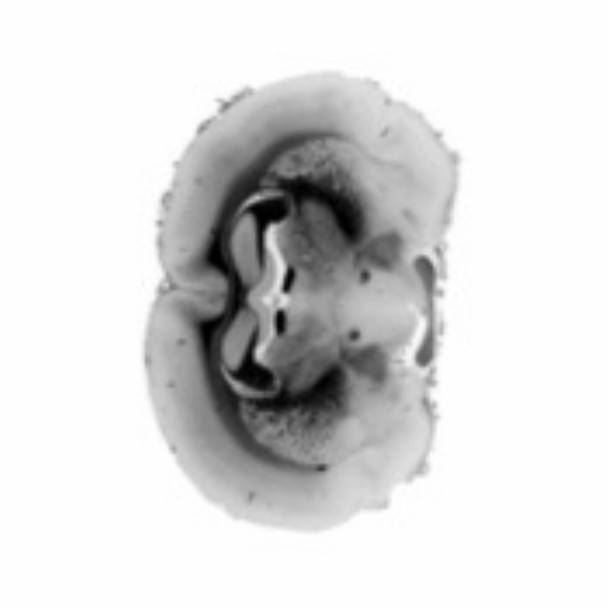}}
%  \centerline  {(b)}\medskip
\end{minipage}
\begin{minipage}[b]{0.32\linewidth}
  \centering
  %TODO:crop the image
  \centerline{\includegraphics[scale=0.085]{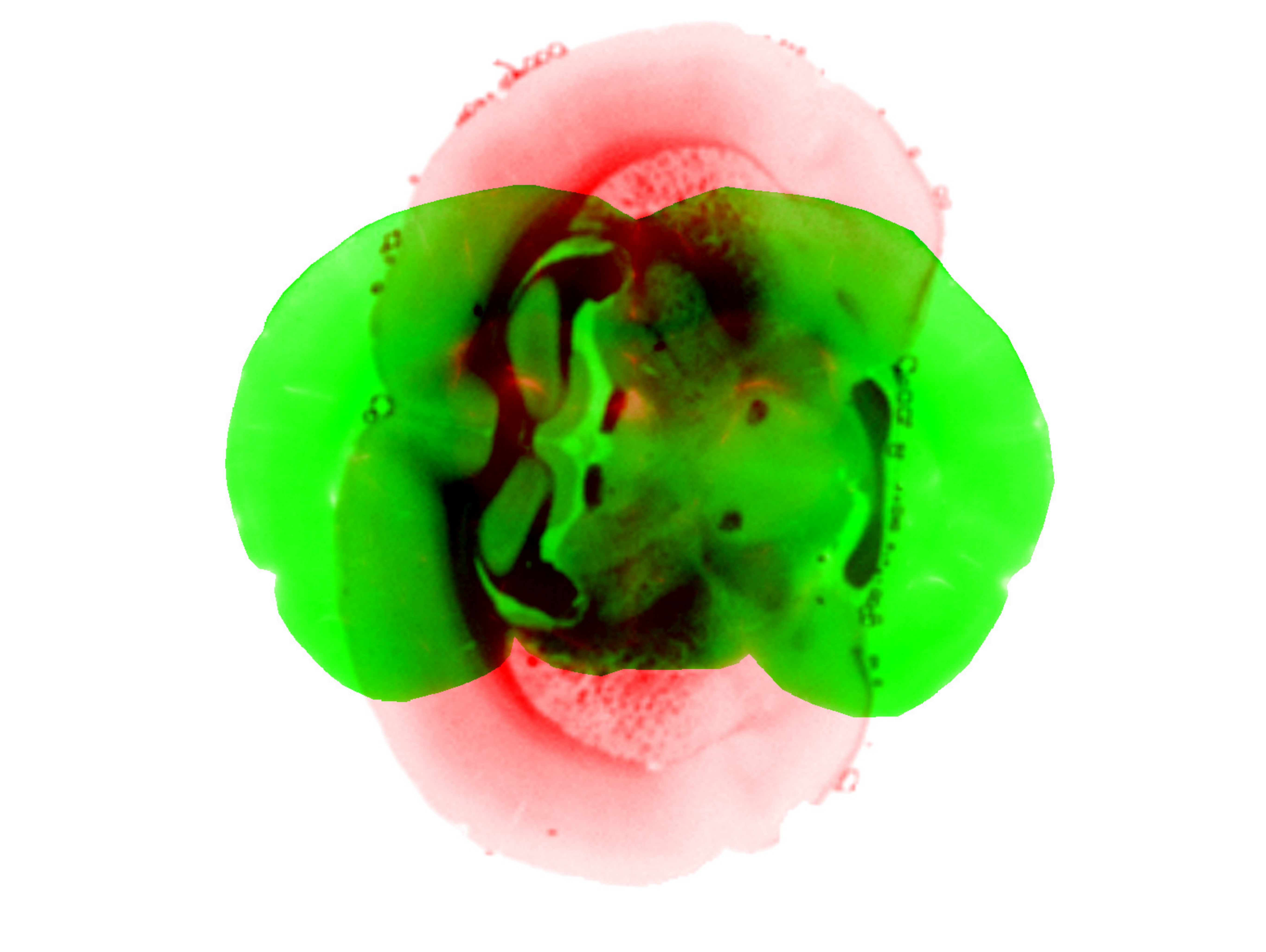}}
 % \centerline  {(c)}\medskip
\end{minipage}
\caption{Pre-processing of high-resolution 3D-PLI data. Left: Original image data, middle: segmented and scaled image, and right: COM alignment with blockface image.}
% using the center-of-mass and isotropic scaling
\label{fig:segmentation}
\end{figure}
%%%%%%%%%
\section{Method}\label{sec:method}
%%%%%%%%%
Our approach for 3D reconstruction of both high- and ultra-high resolution 3D-PLI data of the rat brain consists of several steps. High resolution images are first registered to their corresponding reference blockface images using rigid and non-rigid registration. Ultra-high resolution images are then registered both rigidly and non-rigidly to the corresponding sections of the reference blockface images. %Such transformation of different image resolutions to the reference blockface preserves the brain topology.
\subsection{Registration of high-resolution 3D-PLI data}{\label{HR-reg}}
To coherently align the high-resolution 3D-PLI data with the reference blockface images several registration steps are required. The high-resolution data is first coarsely registered using center-of-mass alignment, rigid registration, and then non-rigid registration using GEBS~\cite{KohlrauschEt05} in conjunction with a novel feature transform-based similarity measure and a weighted quadratic regularization.
\subsubsection{Data preparation and coarse registration}
%%%%%%%%%%%%%%%%%%%%%
High resolution 3D-PLI sections of the rat brain are segmented from the original image data~(see Fig. \ref{fig:segmentation}, left) as in \cite{Ali1Et17:ISBI}. For initial alignment we perform a scaling transformation for high-resolution images ($64~\mu$m) and then align their center-of-mass (COM) with that of the reference blockface images ($15.5~\mu$m, see Fig.~\ref{fig:segmentation}, right). %This is because blockface images used in our experiments have a higher spatial resolution ($15.5\mu$m) compared to that of the high-resolution 3D-PLI data ($64\mu$m). 
%
%\subsubsection{Rigid registration}

We use a parametric registration model for coarse registration of 3D-PLI data. Let $g_{1}(\textbf{x})$ and $g_{2}(\textbf{x})$ with $\textbf{x} = (x, y):\Omega \rightarrow \mathbb{R}, \Omega \in \mathbb{R}^2$, be the reference blockface and the PLI image, respectively, and $\mathcal{T}(\textbf{x}\mid {\theta})$ be the transformation with the parameter vector $\theta$ to be estimated. Then, the goal is to minimize the objective function $\psi$ to obtain the optimal $\hat{\theta}$:
\vspace{-0.15cm}
\begin{equation}{\label{eq:minimize_1}}
~~~~~~\quad \quad\hat{{\theta}} = \argmin_{{\theta}} \psi ~\big(~{{g}_{{1}}{(\textbf{x}), ~ {g}_{2} \big( \mathcal{T}(\textbf{x}\mid {\theta})~\big)}}.
\vspace{-0.25cm}
\end{equation}
We use a spline-based multi-resolution scheme for rigid registration based on~\cite{ThevenazEt98:TIP}. In contrast to~\cite{ThevenazEt98:TIP}, where the sum of squared intensity differences (SSD) was employed, we propose using a correlation transform (CoT) of the image to deal with multi-modal data (see Fig.~\ref{fig:data}). Let $P_{\bf{x}}$ be a patch of size $7\times 7$ pixels centered at $\bf{x}$, then the CoT is given by
\begin{equation}{\label{eq:SSD-CoT}}
~~~~~~\quad \quad{\tilde{g}}\left(\textbf{x}\right) = \left({g}\left(\textbf{x}_k\right) - \mu\right) / (\sigma + \epsilon), \quad  \mathrm{with} ~~\textbf{x}_k \in P_{\bf{x}},
\end{equation}
where $\mu$ and $\sigma$ are the mean intensity and standard deviation, respectively within $P_{\bf{x}}$ and $\epsilon = 0.001$. %{\color{red}CoT is partially invariant to local intensity variations in 3D-PLI data (see Fig. 3, $\tilde{g}$) but can be efficiently computed, and is sufficient for robust global alignment.
%write about CoT and refer to figure
%Since CoTs compute for each pixel a value based on the mean and standard deviation of the image intensities within a local patch, CoTs are invariant to local intensity variations
For $\psi$ in (\ref{eq:minimize_1}) we use the SSD between the computed CoT values for the blockface image $\tilde{g}_{\footnotesize{1}}$ and the high-resolution image $\tilde{g}_{2}$: $\psi(\theta) = \sum_{\textbf{x}\in \Omega} {\Big({\tilde{g}_{{1}}{(\textbf{x})}- \tilde{g}_{2} \big( \mathcal{T}(\textbf{x}\mid {\theta})\big)}\Big)^{2}}.$
We minimize Eq.~(\ref{eq:minimize_1}) using Levenberg-Marquardt optimization.
\vspace{-0.25cm}
% \quad ~ \! \! \!
\begin{figure}[t!]
\begin{minipage}[b]{0.24\linewidth}
  \centering
  %TODO:crop the image
  \centerline{\includegraphics[trim=0.7cm 0.75cm 0.5cm 0.5cm, clip=true, scale=0.5]{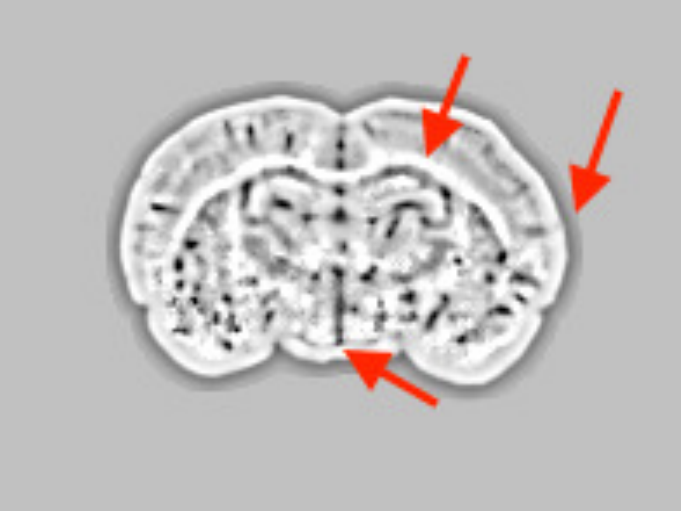}}
 \centerline  {$\tilde{g}_{1}$}\medskip 
\end{minipage}
\begin{minipage}[b]{0.24\linewidth}
  \centering
  %TODO:crop the image
  \centerline{\includegraphics[trim=0.7cm 0.75cm 0.5cm 0.5cm, clip=true, scale=0.5]{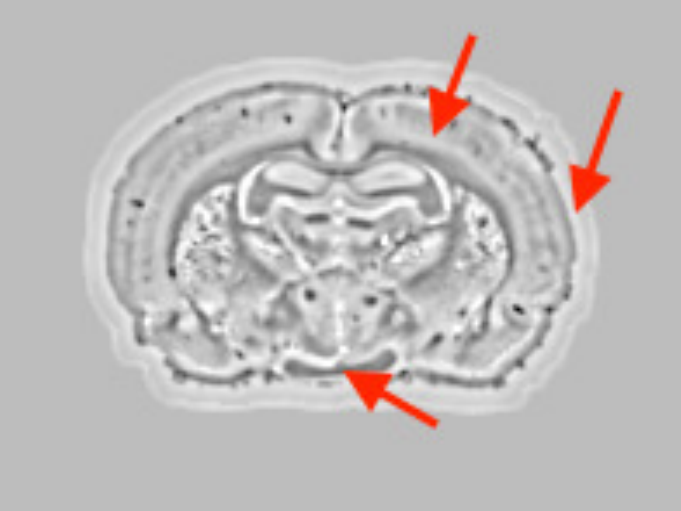}}
  \centerline  {$\tilde{g}_{2}$}\medskip
\end{minipage}
\begin{minipage}[b]{0.24\linewidth}
  \centering
  %TODO:crop the image
  \centerline{\includegraphics[trim=55cm 60cm 60.0cm 35cm, clip=true, scale=0.0073]{s0189_target_FeT-eps-converted-to.pdf}}
 \centerline  {${FeT}_{g_1}$}\medskip
\end{minipage}
\begin{minipage}[b]{0.24\linewidth}
  \centering
  %TODO:crop the image
  \centerline{\includegraphics[trim=55cm 60cm 60.0cm 35cm, clip=true, scale=0.0073]{s0189_source_FeT-eps-converted-to.pdf}}
 \centerline  {${FeT}_{g_2}$}\medskip
\end{minipage}
\caption{Correlation transform ($\tilde{g}$) and feature-transform ($FeT$) images of the multi-modal reference blockface (${g}_1$) and 3D-PLI (${g}_2$) data (also refer to Fig. 1).}
\label{fig:registrationHR}
\end{figure}
\subsubsection{Non-rigid registration}{\label{sec:non-rigid}}
Non-linear distortions are often present in 3D-PLI data due to the cutting and mounting procedure. In addition, local deformations are introduced because of time delays between mounting and data acquisition. Since these deformations are the result of physical phenomena, a suitable physical deformation model should be used for non-rigid registration. In our approach, we use Gaussian elastic body splines (GEBS) which represent an analytic solution of the Navier equation from linear elasticity theory~\cite{ChouPagano67}:
%\begin{equation}{\label{eq:Navier}}
$\mu\Delta\textbf{u}+(\lambda+\mu) \, \nabla\left(\textrm{div}\,\textbf{u}\right)+\textbf{f} = \textbf{0}$,
%\end{equation} 
where $\lambda$ and $\mu$ $>0$ are the Lam{\'e} constants and $\textbf{u}$ is the deformation field under Gaussian forces $\textbf{f}$, and which has been derived in~\cite{KohlrauschEt05}. In~\cite{WoerzRohr14:IJCV}, an intensity-based registration approach using GEBS  was described, which, however is not suitable for multi-modal 3D-PLI data. Using a CoT-based similarity measure for non-rigid registration has disadvantages (see the red arrows in Fig. 3 which indicate that structure and intensity invariance are not well preserved). In this contribution, we introduce a feature transform-based (FeT) similarity measure, and a Gaussian weighted quadratic regularization. FeT better preserves the structure and intensity invariance and is thus better suited for non-rigid registration. FeT consists of: 1) a structure variability measure $S_{var}$ defined by the trace of a covariance matrix $\mathcal{C}$ for seven features: Position ($x$, $y$), absolute values of first and second order image derivatives ($\mid\!\! g_{x}\!\!\mid$, $\mid \!\!g_y\! \!\mid$, $\mid\!\! g_{xx}\!\!\mid$, $\mid\!\! g_{yy}\!\! \mid$), and intensity difference $\mid \! \! g({\bf{x}})\! -\!g({\bf{x}}_k)\! \!\mid$ for each ${\bf{x}}_k$ within the patch $P_{\bf{x}}$, and 2) a texture measure $S_T$ based on cross-correlation between the pixels in $P_{\bf{x}}$ ($5\times5$ pixels). The combined feature transform (FeT) is then designed as a weighted sum of the two components $FeT = S_{var} + 0.5S_T$. Fig. 3 (right) shows example results for $FeT$ for blockface ($FeT_{g_1}$) and PLI images ($FeT_{g_2}$). It can be seen that boundaries and inner texture are quite similar for the multi-modal images. To preserve discontinuities of the deformation field, we use Gaussian weights $f_{\sigma}$ for the quadratic regularization. We use the energy functional
{\small
\begin{eqnarray} {\label{eq:minimization}}
 \argmin_{\bm{u},\bm{u}^I} \sum_\Omega \underbrace{{J_{data}(FeT_{g_1},FeT_{g_2},\bm{u}^{I})
  +\lambda_{I}f_{\sigma}(\parallel \!\!\textbf{u}\!\!\parallel) \parallel \!\!\bm{u}-\bm{u}^I\!\!\parallel_2^2 }}_{J_{Intensity}}
  + \lambda_{E}\, J_{elastic}\!\left(\bm{u}\right),    
\end{eqnarray}
}where $FeT_{g_1}$ and $FeT_{g_2}$ are the feature transforms of the target and source images, respectively. The weighting factors~$\lambda_I, \lambda_E>0$ control the trade-off between the data term $J_{data}$ and the two regularization terms (quadratic and elastic). $\bm{u}^I$ is the deformation field obtained by minimizing the SSD between the feature transforms with a weighted quadratic regularization (i.e. minimization of $J_{Intensity}$) using Levenberg-Marquardt optimization. The final deformation field $\bm{u}$ is obtained using an analytic solution based on GEBS.
\begin{figure}[t!]
\begin{minipage}[b]{0.33\linewidth}
  \centering
  %TODO:crop the image
  \centerline{\includegraphics[trim=5cm 5cm 2.0cm 2cm, clip=true, scale=0.14]{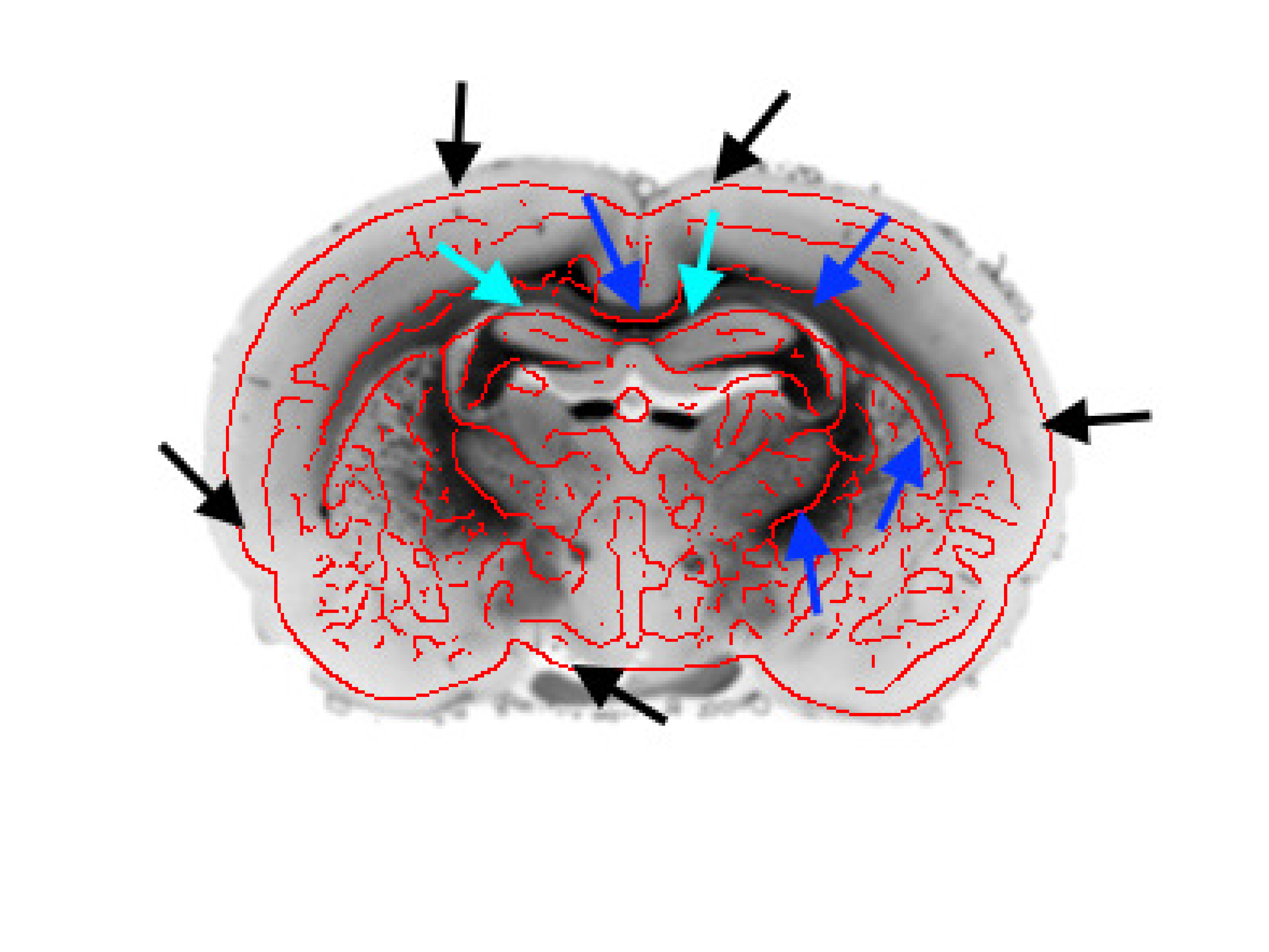}}
 % \centerline  {(a)}\medskip
\end{minipage}
\begin{minipage}[b]{0.33\linewidth}
  \centering
  %TODO:crop the image
  \centerline{\includegraphics[trim=5cm 5cm 3.0cm 2cm, clip=true, scale=0.14]{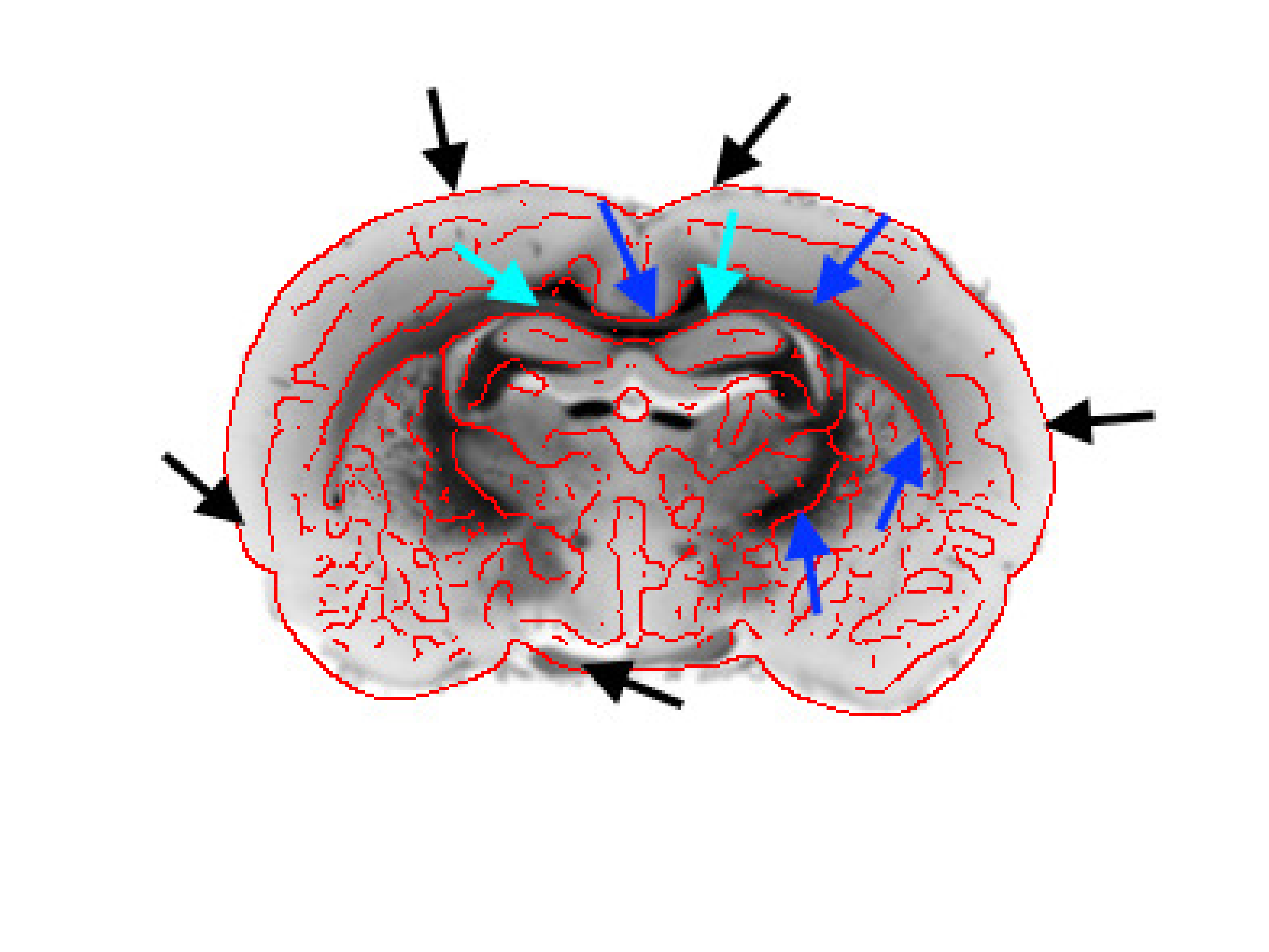}}
%  \centerline  {(b)}\medskip
\end{minipage}
\begin{minipage}[b]{0.32\linewidth}
  \centering
  %TODO:crop the image
  \centerline{\includegraphics[trim=0cm 0cm 0cm 0cm, clip=true, scale=0.21]{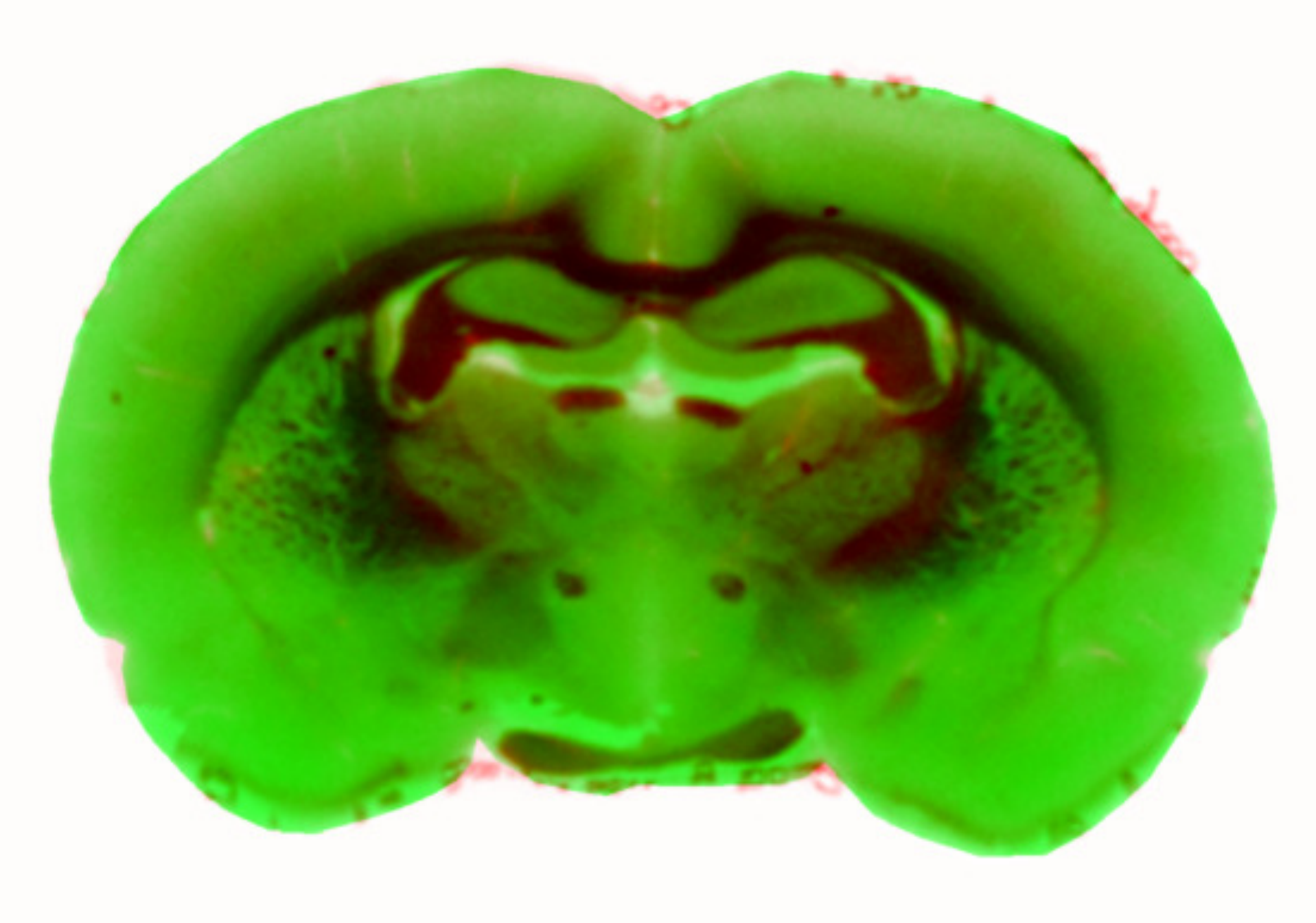}}
 % \centerline  {(c)}\medskip
\end{minipage}
\caption{Registration of high-resolution 3D-PLI data. Left: Rigid  registration, middle: non-rigid registration (edges of blockface overlaid with high-resolution 3D-PLI image), and right: Color-overlay image with blockface (green) and registered high-resolution3D-PLI  image (red).}
\label{fig:registrationHR}
\end{figure}

Fig.~\ref{fig:registrationHR} (left) reveals the result after rigid registration. Visual inspection shows a good alignment, however, misalignments are distinct along the corpus callosum (indicated by blue arrows), the hippocampus (cyan arrows), and along the borders of the cerebral cortex (black arrows).
%%%%%
Using the new similarity measure for non-rigid registration, it can be observed in Fig.~\ref{fig:registrationHR} that the misalignments in various regions have  been tackled (see Fig.~\ref{fig:registrationHR}, middle and right). 
\subsection{Registration of ultra-high resolution 3D-PLI data}
\begin{figure}[t!]
\begin{minipage}[b]{0.24\linewidth}
  \centering
  \centerline{\includegraphics[height=1.7cm, width=3cm]{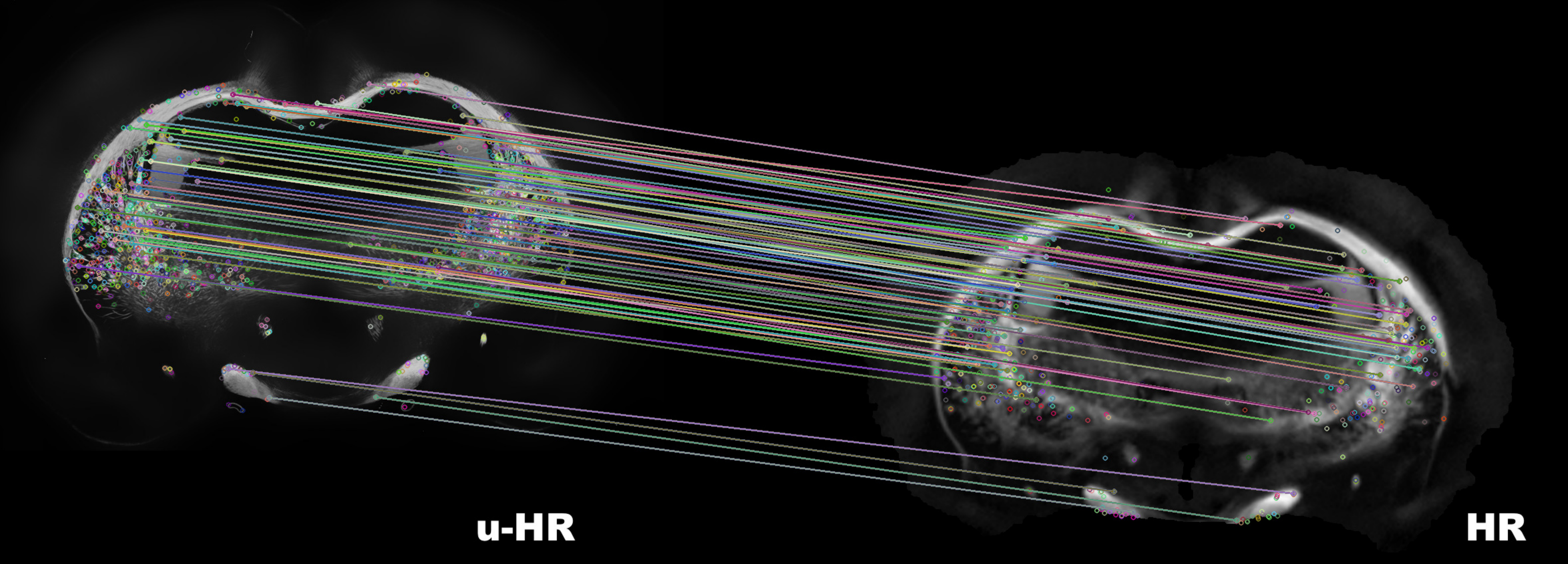}}
%  \centerline  {(a)}\medskip
\end{minipage}
\begin{minipage}[b]{0.24\linewidth}
  \centering
  \centerline{\includegraphics[trim=2cm 1.25cm 2cm 0.5cm, clip=true, scale=0.025]{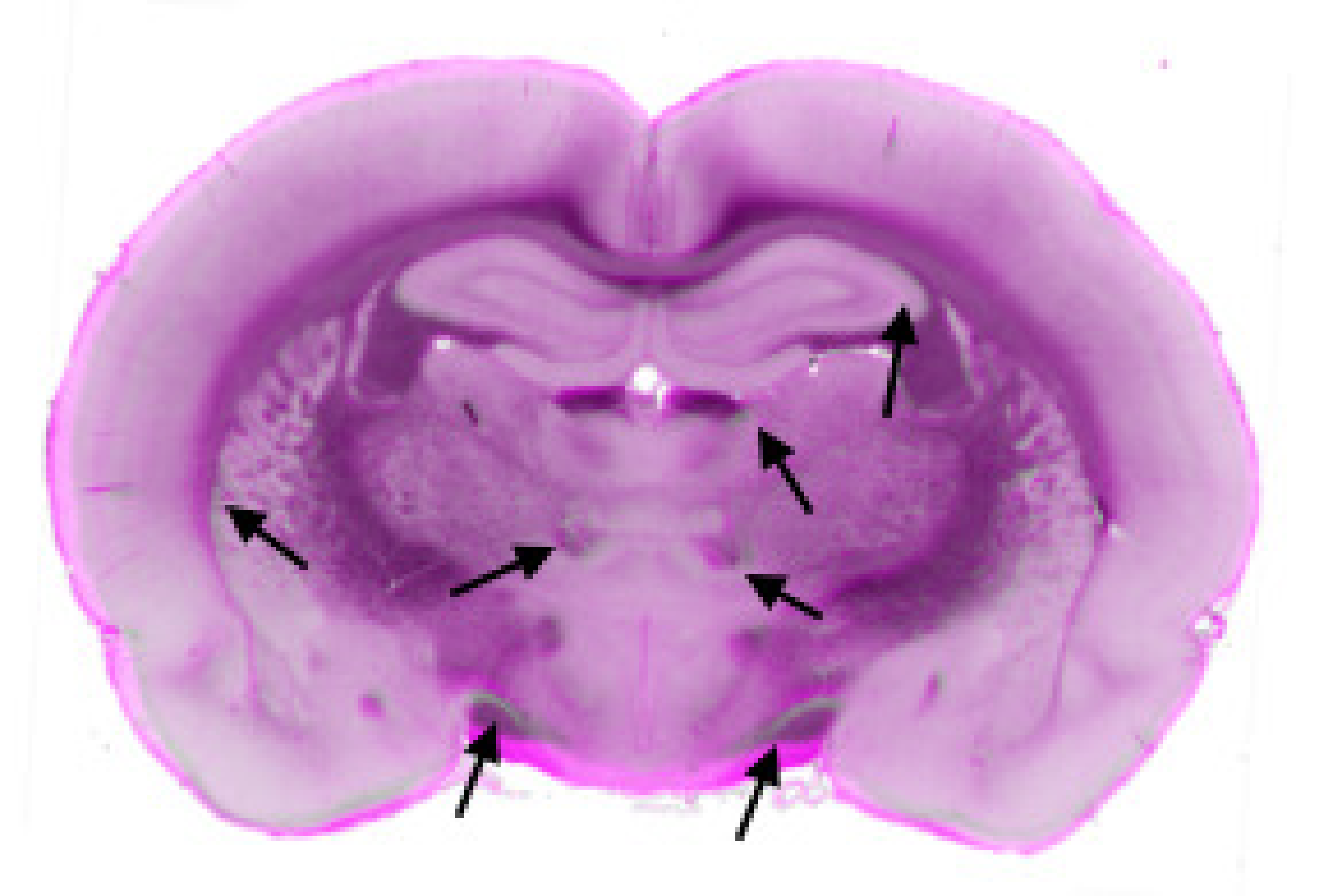}}
%  \centerline  {(b)}\medskip
\end{minipage}
\begin{minipage}[b]{0.24\linewidth}
  \centering
  \centerline{\includegraphics[trim=3cm 1.25cm 2cm 0.5cm, clip=true, scale=0.025]{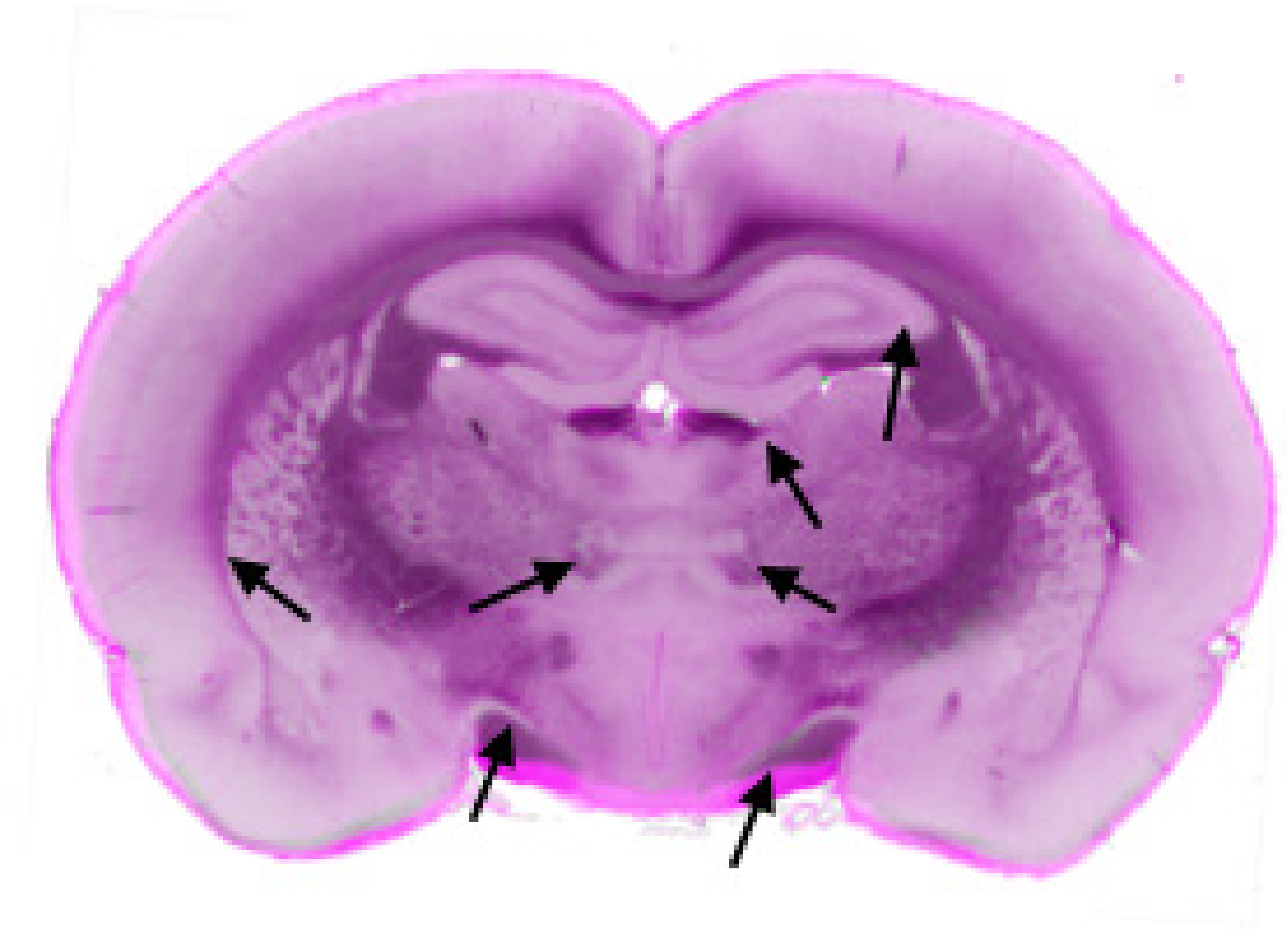}}
%  \centerline  {(b)}\medskip
\end{minipage}
\begin{minipage}[b]{0.24\linewidth}
  \centering
  \centerline{\includegraphics[trim=3cm 1.25cm 2cm 0.5cm, clip=true, scale=0.025]{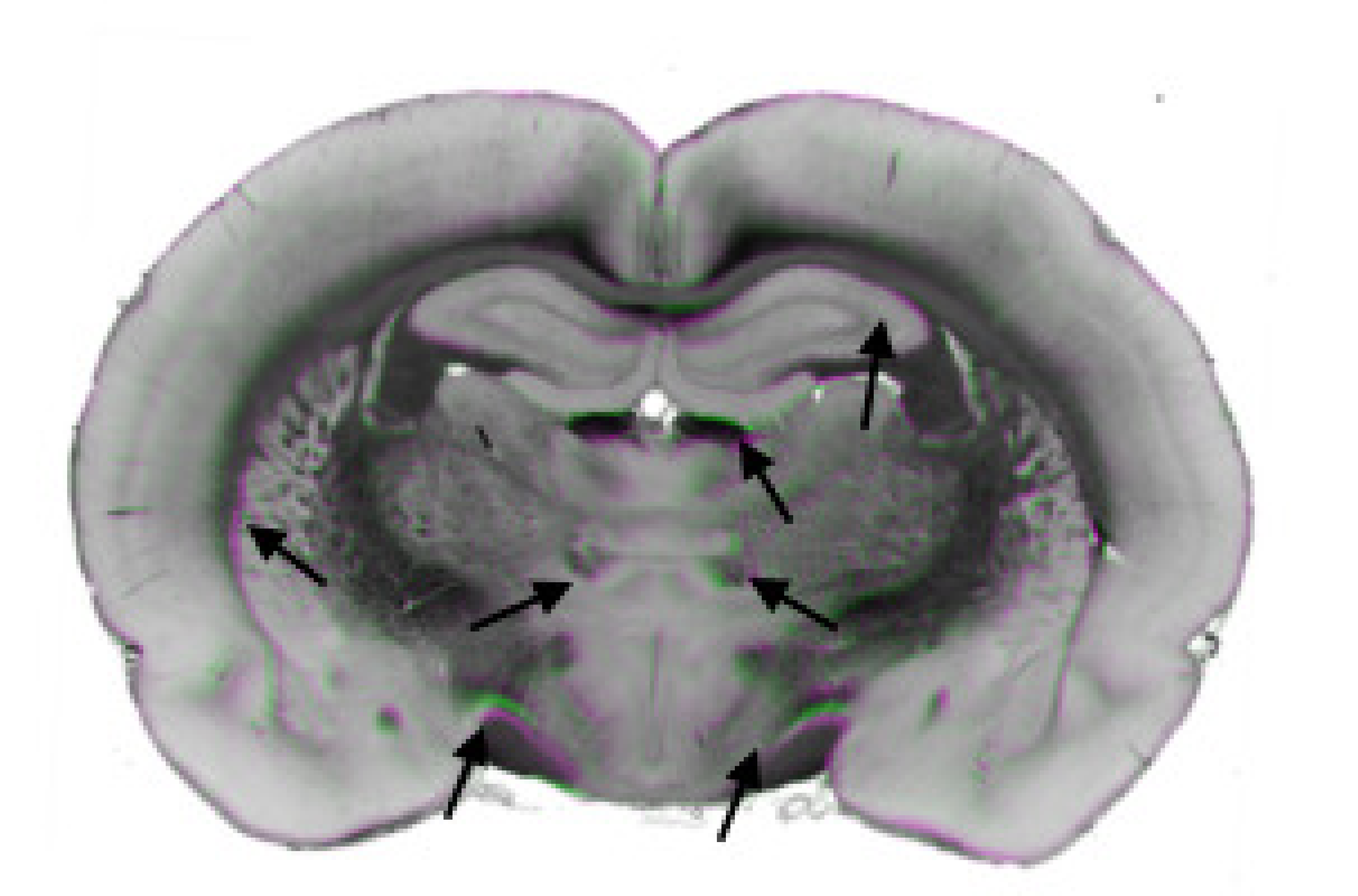}}
%  \centerline  {(b)}\medskip
\end{minipage}
%%%%%%%%%------> 337
\begin{minipage}[b]{0.24\linewidth}
  \centering
  \centerline{\includegraphics[trim=0cm 3.0cm 1cm 0cm, clip=true, height=1.7cm, width=3cm]{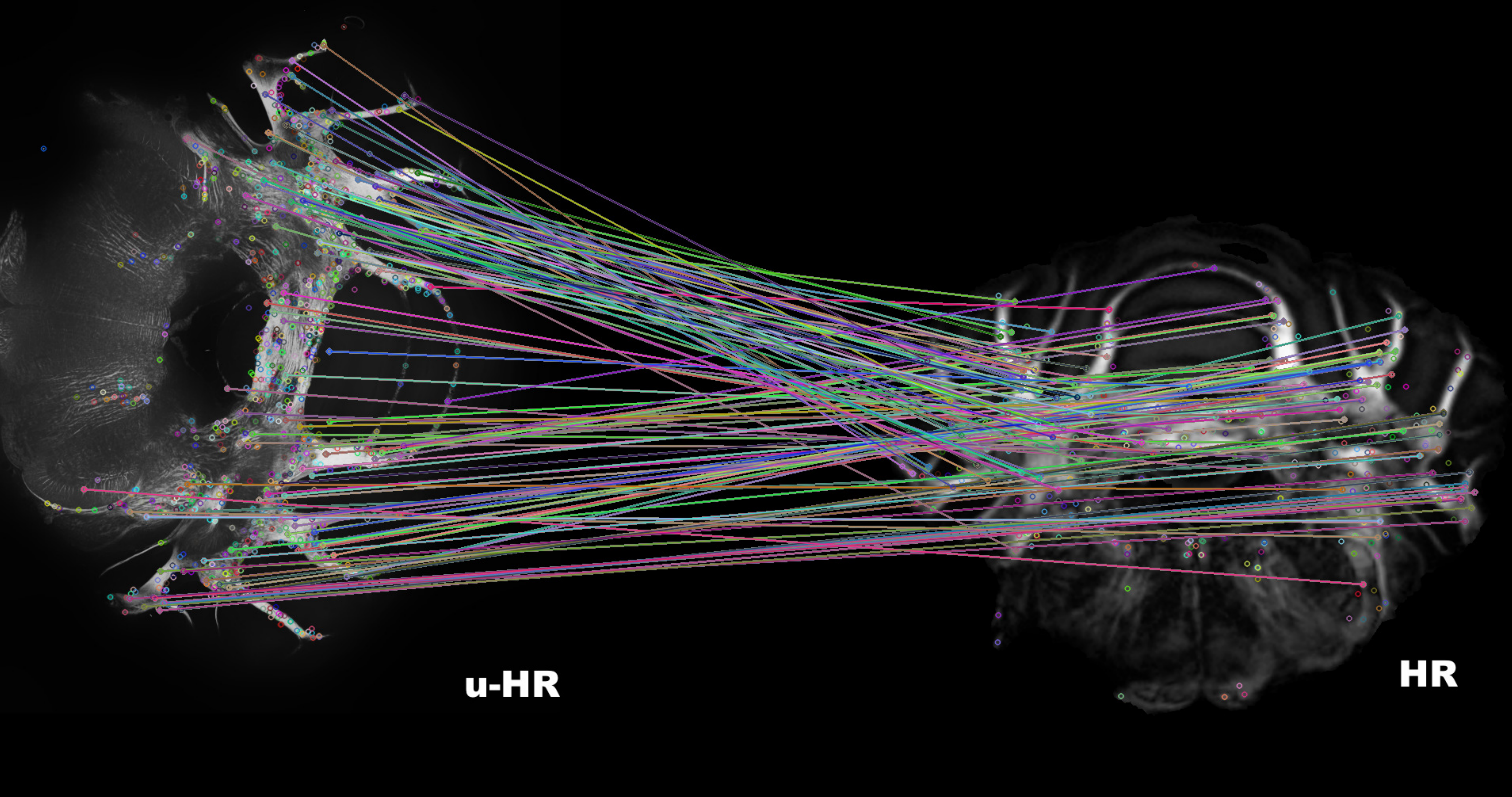}}
% \centerline  {(a)}\medskip
\end{minipage}
\begin{minipage}[b]{0.24\linewidth}
  \centering
  \centerline{\includegraphics[trim=2cm 2.25cm 2cm 0.5cm, clip=true, scale=0.03]{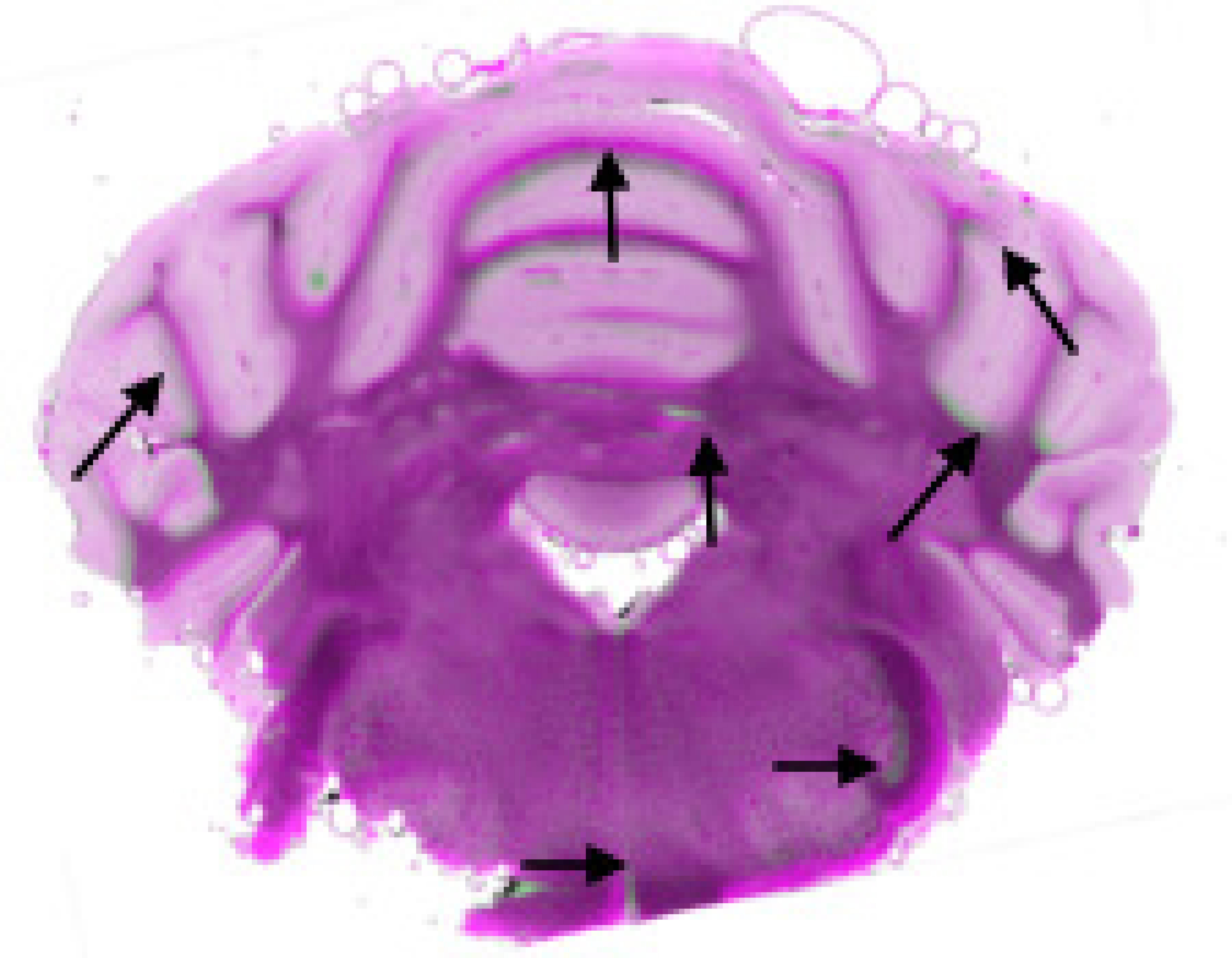}}
%\centerline  {(b)}\medskip
\end{minipage}
\begin{minipage}[b]{0.24\linewidth}
  \centering
  \centerline{\includegraphics[trim=4cm 1.25cm 2cm 0.5cm, clip=true, scale=0.03]{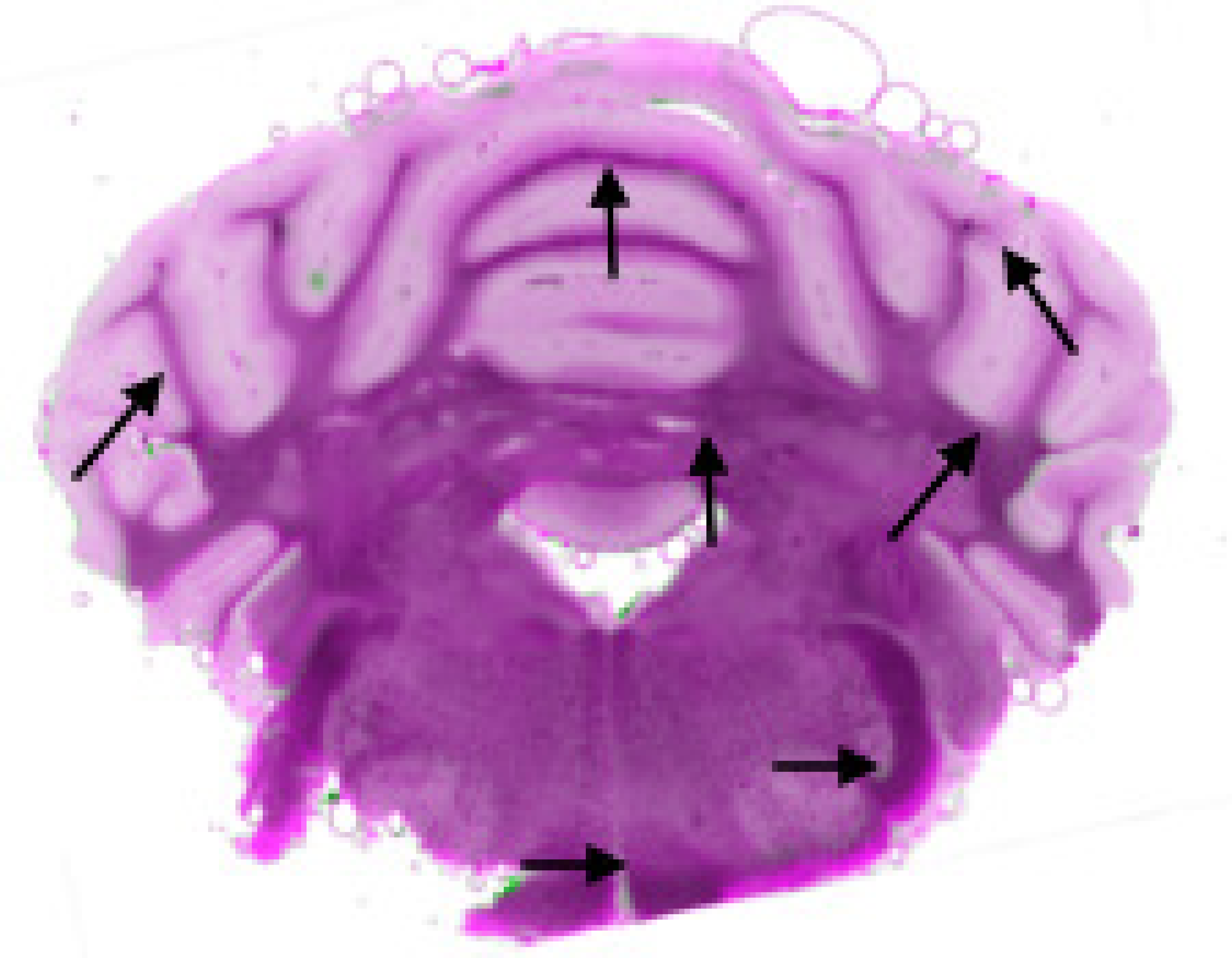}}
 % \centerline  {(c)}\medskip
\end{minipage}
\begin{minipage}[b]{0.24\linewidth}
  \centering
  \centerline{\includegraphics[trim=3cm 1.25cm 2cm 0.5cm, clip=true, scale=0.03]{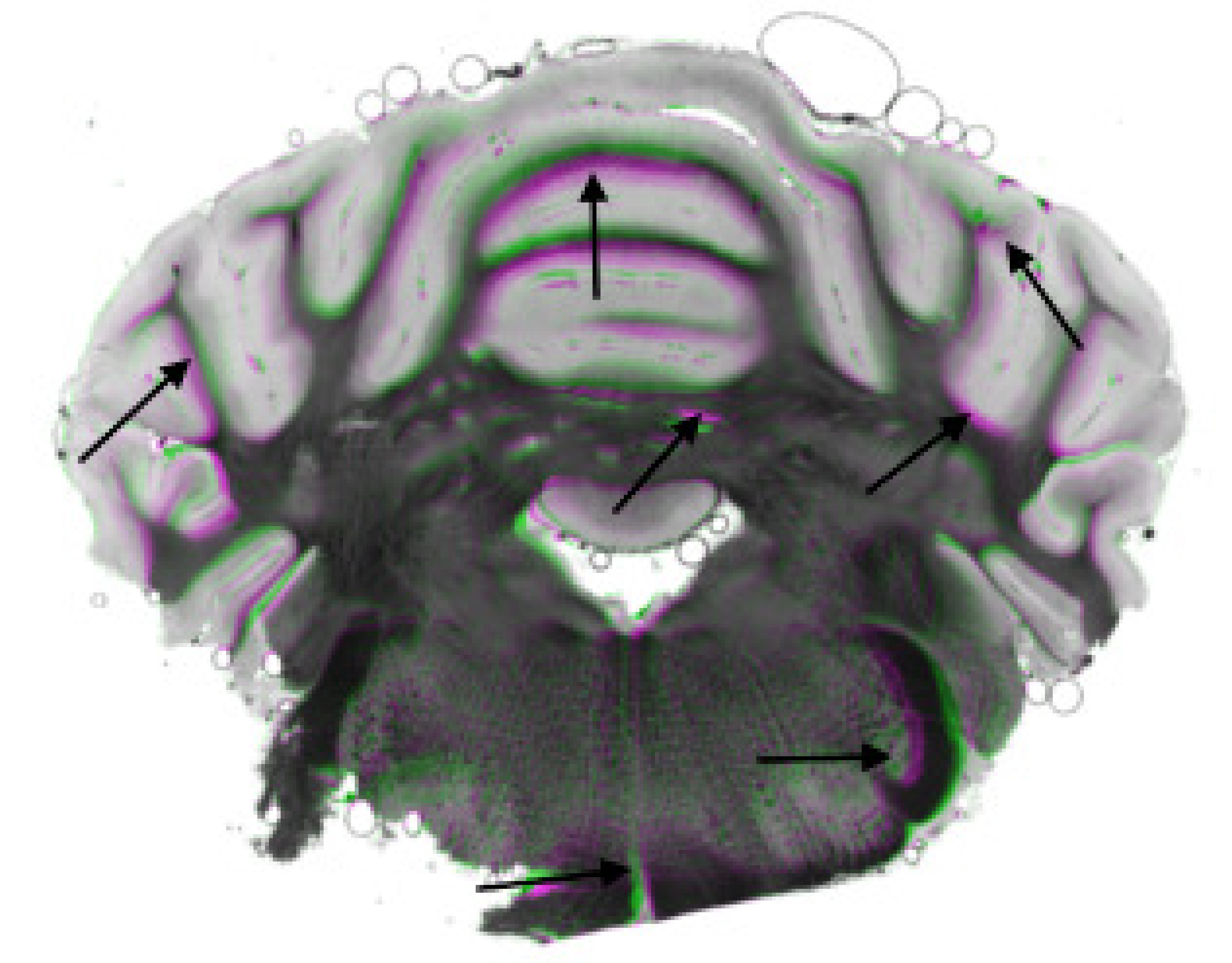}}
% \centerline  {(d)}\medskip
\end{minipage}
\caption{Registration of ultra-high resolution (u-HR) images to the registered high-resolution (HR) images. First column: Matching of detected salient features (retardation maps), second column: Alignment after similarity transformation, third column: Alignment after non-rigid registration (transmittance maps), and fourth column: Highlighted regions with large deformations. Misalignments are indicated by green regions and black arrows.}
\label{fig:registrationuHR1}
\end{figure}
% For 

Due to the large difference in spatial resolution between the blockface images and the ultra-high resolution images (factor of about 12) and arbitrary rotations we perform registration using a scale-space method for feature detection and matching. A Gaussian scale-space and a Hessian measure are used to detect features in the registered high-resolution and the ultra-high resolution (retardation map) images with subsequent feature matching based on FLANN~\cite{Ali2Et17:ISBI}. Then, a similarity transformation (rigid and isotropic scaling) is computed using the matched features and a least squares approach. Unlike in~\cite{Ali2Et17:ISBI}, we use a fast bilateral filtering technique~\cite{ParisDurand09:IJCV} to cope with the noise in the rat brain data and reduce false detections in feature extraction. In Fig.~\ref{fig:registrationuHR1}, examples for feature matching results are shown. Subsequently, a non-rigid registration (see Sect.~\ref{sec:non-rigid}) is used to cope with local deformations at $15.5~\mu m$ resolution (see Fig.~\ref{fig:registrationuHR1}, third column). Further, visible misalignments in Fig.~\ref{fig:registrationuHR1} (third column) are corrected at a resolution of $1.3~\mu m$ using the proposed non-rigid registration method (see Eq. (\ref{eq:minimization})) and coarse-to-fine energy minimization (we use 9 pyramid levels).
%the first column shows the matched salient feature points using the method proposed.
%~For $\sigma_s$ and ~$\sigma_r$ we used the proposed values in~\cite{ParisDurand09:IJCV}. This allows the preservation of only salient features while suppressing amplified noise in 3D-PLI images. 
%
% Illustrate with the examples and images here
%Due to the time interval between imaging of the high-resolution data and the ultra-high-resolution data some local deformations may exist. To cope with these deformations we use the non-rigid registration method. Due to  variations in pixel intensities and noise in these two modalities we use our novel local metric introduced in Section~\ref{sec:non-rigid}. It is evident in the second column of Fig.~\ref{fig:registrationuHR1} that such misalignments are present which is corrected using the proposed non-rigid approach (see third column of Fig.~\ref{fig:registrationuHR1}).
%
\section{Experimental results}\label{sec:results}
We have evaluated the proposed method for the registration and 3D reconstruction of high- and ultra-high-resolution data of the rat brain (64 $\mu m$ and 1.3 $\mu m$). Ground truth correspondences for three sections were determined manually by an expert (on average 25 and 46 landmarks for high- and ultra-high resolution sections, respectively). Table \ref{table:LMevaluation} shows the average target registration error (TRE). It can be seen that our proposed non-rigid registration method using the feature transform $FeT$ yielded an overall improvement of about $4.1$ pixels and $4.8$ pixels compared to a previous non-rigid registration approach using mutual information~\cite{BiesdorfEt09:MICCAI} for high- and ultra-high resolution 3D-PLI data, respectively. Notably, our non-rigid registration method can deal with large deformations which is evident from the large overall improvements of 15.7 pixels and 10.9 pixels compared to rigid registration using CoT ($\tilde{g}$) for high-resolution and ultra-high resolution data, respectively. Rigid registration using the original image data ($g$) and center-of-mass alignment (COM) yielded worse results.
\begin{table}[t!]
\centering
%\small{
\centering
\begin{tabular}{|c||cc||cccccc||cccc|}
\hline
\multicolumn{1}{|c|}{\multirow{2}{*}{\textbf{Section}}} &\multicolumn{2}{|c|}{}& \multicolumn{6}{|c|}{\textbf{HR}($64\mu m$) $\Rightarrow$ \textbf{BF}} & \multicolumn{4}{|c|}{\textbf{u-HR}($1.3 \mu m$) $\Rightarrow$ \textbf{BF}}  \\ \cline{4-8} \cline{8-12}
\multicolumn{1}{|c|}{}  & \multicolumn{2}{c|}{\textbf{LMs}}&  ~initial &\multicolumn{2}{c}{~rigid} &       &\multicolumn{2}{c|}{~~non-rigid}  & initial &  ~~rigid  &  \multicolumn{2}{c|}{~~non-rigid}    \\ \cline{5-6} \cline{8-9}\cline{12-13}
\multicolumn{1}{|c|}{}  &  \multicolumn{1}{c|}{HR}& \multicolumn{1}{c|}{u-HR}&\multicolumn{1}{|c}{COM}   & \multicolumn{1}{c|}{$g$} & \multicolumn{1}{c}{$\tilde{g}$}   &  & \multicolumn{1}{c|}{MI}    &  \multicolumn{1}{c|}{$FeT$} &  \multicolumn{1}{|c}{COM} &   + scale &  \multicolumn{1}{c|}{MI}  &   \multicolumn{1}{c|}{$FeT$} \\ 
\hline \hline
%-----------------------------------------------------------------------------------------%		                           
% this is for PM
\# 105 		& \multicolumn{1}{c|}{25}& 35 	&708.3 & \multicolumn{1}{c|}{37.7} & 23.9  & &    \multicolumn{1}{c|}{\bf{7.0}} &7.2
	% for PM
	 & 419.7  & 8.4 &  13.5 & \bf{4.9}\\
			&	\multicolumn{1}{c|}{}& 	&$\pm$\scriptsize{93.4} & \multicolumn{1}{c|}{$\pm$\scriptsize{16.8}} &$\pm$\scriptsize{14.7} &  & \multicolumn{1}{c|}{$\pm$\scriptsize{6.7}}  &  $\pm$\scriptsize{\bf{3.2}}& $\pm$\scriptsize{13.4}& $\pm$\scriptsize{3.5}&$\pm${{\scriptsize{6.8}}}    &$\pm${\bf{\scriptsize{2.9}}} \\  \hline
%-----------------------------------------------------------------------------------------%		                          
% this is for PM
\# 131		& \multicolumn{1}{c|}{22}&42 	&785.8 &  \multicolumn{1}{c|}{40.3} & 22.2 & &    \multicolumn{1}{c|}{13.9} & \bf{6.0}
	% for PM
	& 532.7  & 25.7 &13.3 & \bf{10.4} \\
	&	\multicolumn{1}{c|}{}		& 	 &$\pm$\scriptsize{247.5}  &\multicolumn{1}{c|}{$\pm$\scriptsize{10.6}} & $\pm$\scriptsize{6.8}&  &\multicolumn{1}{c|}{$\pm$\scriptsize{3.5}} & $\pm$\scriptsize{\bf{3.3}} & $\pm$\scriptsize{271.0}&$\pm$\scriptsize{9.9} & $\pm${\bf{\scriptsize{5.7}}} & $\pm${\scriptsize{5.9}} \\  \hline
%-----------------------------------------------------------------------------------------%		                        
% this is for PM
%\# 161		&\multicolumn{1}{c|}{27}&42 	& 638.7 &  \multicolumn{1}{c|}{41.7} & 12.7&  & \multicolumn{1}{c|}{5.3} & 5.9
%	% for PM
%	& 628.0  & 12.8  & &11.4   \\
%		&	\multicolumn{1}{c|}{}	&  &$\pm$230.3 &\multicolumn{1}{c|}{$\pm$11.1} & $\pm$6.5&  & \multicolumn{1}{c|}{$\pm$ 3.7} & $\pm$ 4.1&$\pm$299.8 &$\pm$8.1 &   &$\pm$ 8.1\\ \hline
%-----------------------------------------------------------------------------------------%		                    
% this is for PM
%\# 196 		& \multicolumn{1}{c|}{23}&37 	&723.6 & \multicolumn{1}{c|}{44.8}&13.1  &   \multicolumn{1}{c|}{7.8} & 6.8 
%	% for PM
%	& 343.3  & 8.8	&  &  7.6 \\
%	&	\multicolumn{1}{c|}{}		& 	 & $\pm$150.5&\multicolumn{1}{c|}{$\pm$9.2}& $\pm$4.9&  \multicolumn{1}{c|}{$\pm$5.3} &$\pm$2.6  & $\pm$15.7&$\pm$4.8 & & $\pm$4.4\\ \hline
%-----------------------------------------------------------------------------------------%	\hline                           
% this is for PM
\# 337 		& \multicolumn{1}{c|}{29}&61 	&701.6 & \multicolumn{1}{c|}{25.0 }& 20.9&  &    \multicolumn{1}{c|}{11.4}& \bf{6.7}  
	% for PM
	& 453.1  & 21.6 &10.9  &  \bf{7.9} \\
	&		\multicolumn{1}{c|}{}	& 	 & $\pm$\scriptsize{183.7}& \multicolumn{1}{c|}{$\pm$ \scriptsize{8.1}}&$\pm$\scriptsize{7.3} &  &\multicolumn{1}{c|}{$\pm$\scriptsize{11.0}} &$\pm$\scriptsize{\bf{3.3}} &$\pm$\scriptsize{264.6}& $\pm$\scriptsize{7.7}& $\pm${{\scriptsize{9.0}}}& $\pm${\bf{\scriptsize{4.5}}} \\ \hline\hline
%-----------------------------------------------------------------------------------------%										
%%%MEAN values				                       
% this is for PM
	\textbf{Mean: } & \multicolumn{1}{c|}{25}&46 	&731.9 & \multicolumn{1}{c|}{34.3}& 22.3 &     &\multicolumn{1}{c|}{10.7} & \bf{6.6}
	% for PM
	& 475.4  & 18.6 & 12.5 & \bf{7.7} \\
	&	\multicolumn{1}{c|}{}		& 	 & $\pm$\scriptsize{174.8} &  \multicolumn{1}{c|}{$\pm$\scriptsize{11.8}}& $\pm$\scriptsize{9.6} & & \multicolumn{1}{c|}{$\pm$\scriptsize{7.0}} &  $\pm$\scriptsize{\bf{3.3}}& $\pm$\scriptsize{173.0} & $\pm$\scriptsize{7.0}&$\pm$\scriptsize{7.2} & $\pm$\bf{\scriptsize{4.4}}   \\ \hline				
\end{tabular}
\vspace{0.15cm}
\caption{Target registration error and standard deviation ({TREs}$\pm$std. dev.) using landmarks (LMs) from an expert for three rat brain sections. Registration of high-resolution (HR) images and ultra-high-resolution (u-HR) images to reference blockface (15.5 $\mu$m).}{\label{table:LMevaluation}}
\vspace{-0.80cm}
\end{table} 
%
%
%Table \ref{table:LMevaluation} shows the average target registration error (TRE) for COM alignment (after scaling), for rigid, for affine, and for non-rigid registration of high-resolution images to the reference blockface images. Also, TRE for COM alignment (after scaling), for rigid, and non-rigid registration of ultra-high resolution data is presented. 
%
%
%%%%%%%%%%%%%%%%%%%%%
\begin{figure}[b!]
\begin{minipage}[b]{0.32\linewidth}
  \centering
  %TODO:crop the image
%  \centerline{\includegraphics[trim=1.45cm 0cm 0.75cm 1.5cm, clip=true,scale = 0.21]{figs/3D/rigid-cropped.eps}}
  \centerline{\includegraphics[trim=1.45cm 0cm 0.75cm 1.5cm, clip=true,scale = 0.21]{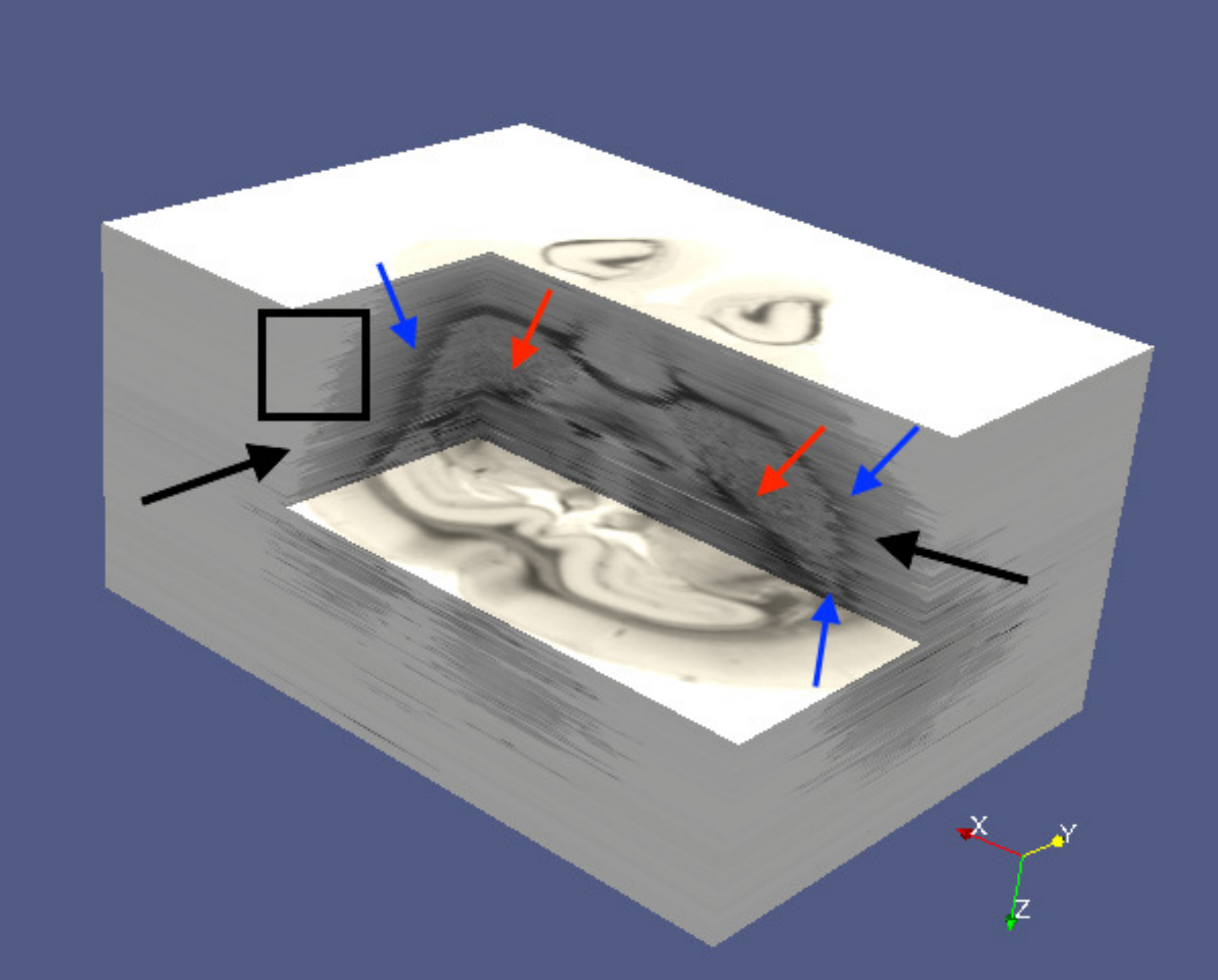}}
 % \centerline  {(a)}\medskip
\end{minipage}
\begin{minipage}[b]{0.32\linewidth}
  \centering
  %TODO:crop the image
%  \centerline{\includegraphics[trim=1cm 0cm 0.25cm 1cm, clip=true,scale=0.188]{figs/3D/d3-cropped.eps}}
 \centerline{\includegraphics[trim=1cm 0cm 1cm 1cm, clip=true,scale=0.188]{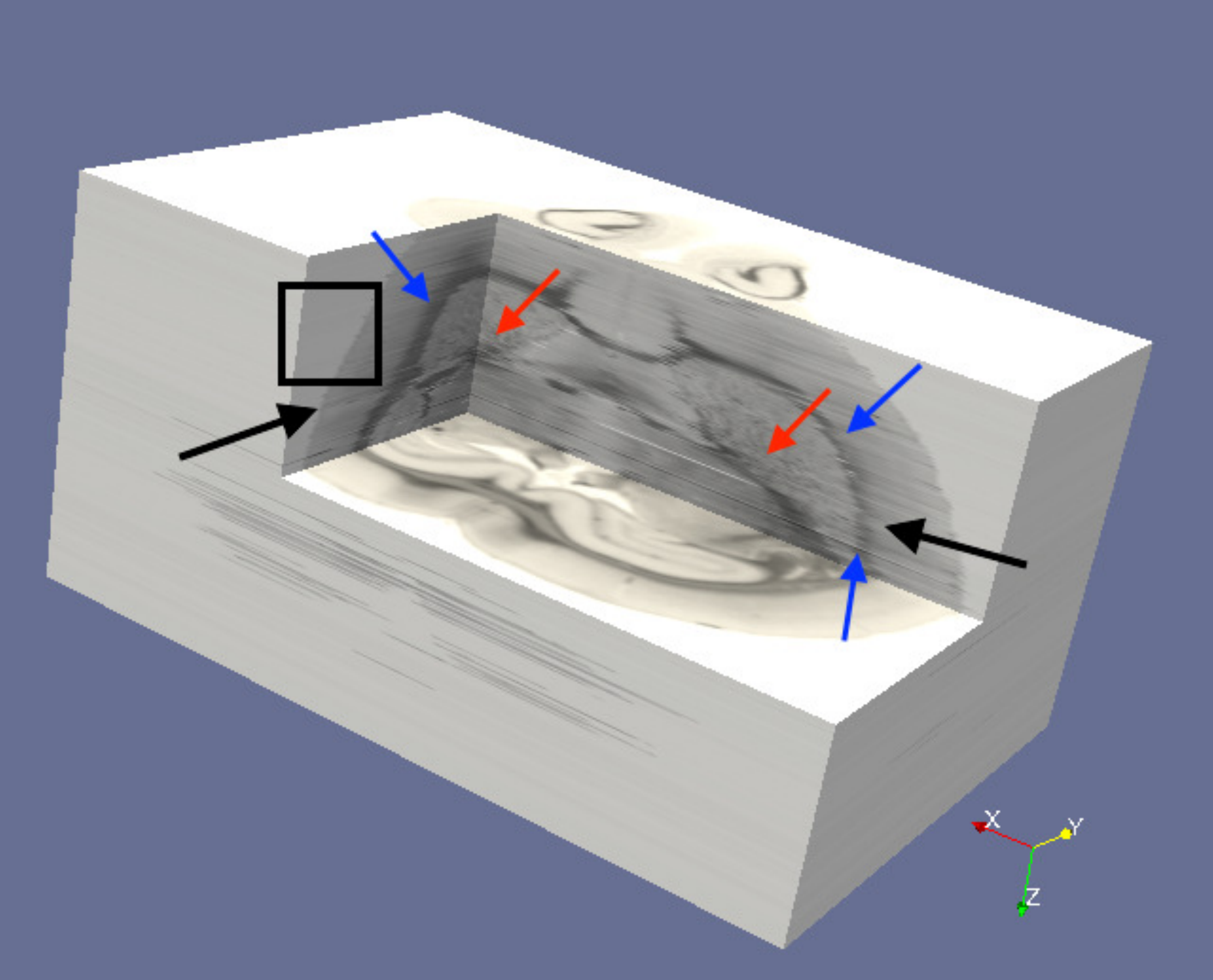}}
%  \centerline  {(b)}\medskip
\end{minipage}
\begin{minipage}[b]{0.33\linewidth}
  \centering
  %TODO:crop the image
%  \centerline{\includegraphics[trim=0.5cm 0cm 0.cm 0.25cm, clip=true,scale=0.31]{figs/3D-1a.eps}}
\centerline{\includegraphics[trim=2.5cm 0cm 0.5cm 0.4cm, clip=true,scale=0.06]{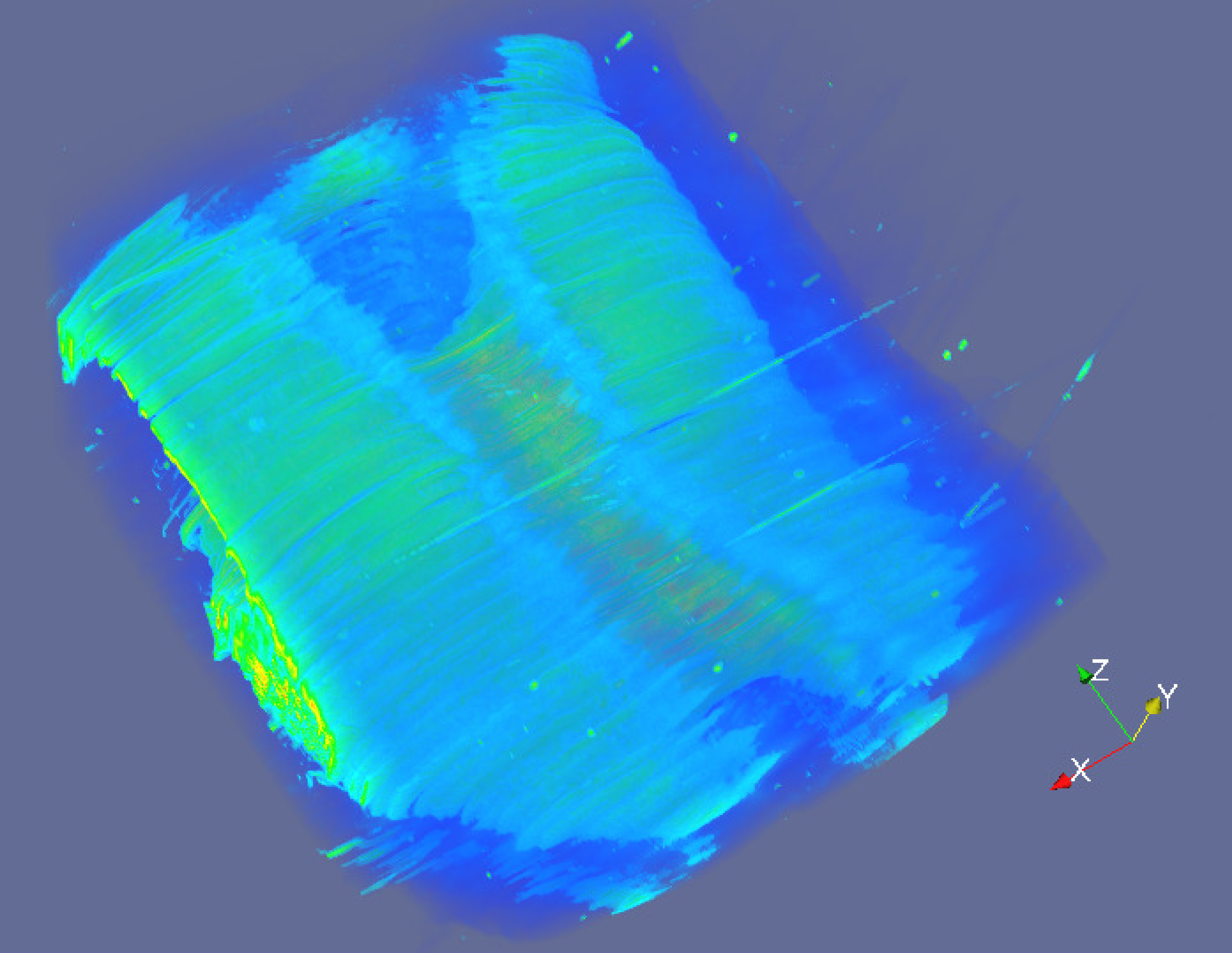}}
%  \centerline  {(b)}\medskip
\end{minipage}
\vspace{0.15cm}
\caption{3D reconstruction. Left: Rigid registration, middle: non-rigid registration, and right: rendered 3D volume at $1.3~\mu m$ resolution (scaled for visualization).}
\label{fig:registrationuHR2}
\end{figure}
%%%%
%
%an improvement compared to mutual information-based (MI) multi-modal measure~\cite{BiesdorfEt09:MICCAI}. 
%
%TRE improvement was more than $4.1$ pixels for three sections, and a mean TRE of $6.4$ with standard deviation of $3.1$ was obtained which is an overall improvement of about $5$ pixels compared to rigid registration.
%Aditionally, non-rigid registration using wNCC can deal with large deformations which is evident from the large improvement compared to affine registration for the brain section \#337 (more than 15.5 pixels). A very large improvement of over 19 pixels is also obtained with the proposed rigid registration method compared to \cite{ThevenazEt98:TIP} ($g$ vs. $\tilde{g}$ in Tab.~\ref{table:LMevaluation}). }

Fig.~\ref{fig:registrationuHR2} (left, middle) shows 3D visualizations of registration results as a reconstructed 3D volume of 278 high-resolution image sections (transmittance maps, $15.5~\mu m \times 15.5~\mu m \times 16.7~mm$). After rigid registration, misalignments are visible at locations indicated by arrows (black: Tissue boundary, blue: Corpus callossum and red: Caudate putamen) and a square in Fig.~\ref{fig:registrationuHR2} (left). However, after non-rigid registration a coherent alignment can be observed (see Fig.~\ref{fig:registrationuHR2}, middle). A rendered 3D reconstructed volume of ultra-high resolution is shown in  Fig.~\ref{fig:registrationuHR2} (right) where the smooth green regions indicate coherent alignment of corpus callossum (retardation maps, $1.3~\mu m \times 1.3 ~\mu m \times 16.7 ~mm$). %For the registration of ultra-high resolution images with blockface images, quantitative results are shown in Tab.~\ref{table:LMevaluation}. It can be seen that our feature-based similarity transformation approach well registers the images (also see second column of Fig.~\ref{fig:registrationuHR1}). Using our proposed non-rigid approach local misalignments are improved. An improvement of 2.2 pixels compared to a similarity transformation (rigid and scale) is obtained.
%, a visual inspection in the third column of Fig.~\ref{fig:registrationuHR1} also validates\emph{\color{red} for significant improvement compared to rigidly registered images shown in second column of Fig.~\ref{fig:registrationuHR1}}.
%
\section{Conclusion}
\label{sec:concl}
We have introduced a new multi-scale and multi-modal registration method for 3D reconstruction of both high-resolution and ultra-high resolution 3D-PLI histological images of a rat brain. The method comprises a novel feature transform-based similarity metric integrated in a physically-based non-rigid registration approach as well as a correlation transform-based similarity measure for robust rigid registration. Quantitative evaluations showed that our method improves the result compared to a previous multi-modal non-rigid registration approach and leads to a coherent 3D reconstruction.
%
%We have presented a novel weighted normalized cross-correlation intensity similarity measure in an non-rigid Gaussian body splines based registration for robust and accurate registration of 3D-PLI data of rat brain. Registration of images with spatial resolution of 64$\mu$m  and $1.3\mu$m images to the reference 15.5$\mu$m blockface are presented. We have validated each of the registration step used for the 3D reconstruction using both quantitative and qualitative evaluations. The future work will include global registration correction in the 3D reconstructed volume.
%
%==================================================================================
\section*{Acknowledgments}
%%==================================================================================
\noindent{}This project was funded by the Helmholtz Association
through the Helmholtz Portfolio theme ``Supercomputing and Modeling for the Human Brain`` and by the European Union through the Horizon 2020 Research and Innovation Programme under Grant Agreement No.~7202070 (Human Brain Project SGA1). 
\bibliographystyle{plain}
\bibliography{registration}
\end{document}